\documentclass{article}

\usepackage{microtype}
\usepackage{graphicx}
\usepackage{subfigure}
\usepackage{booktabs} % for professional tables
\usepackage{hyperref}
\usepackage{url}
\usepackage{amsmath,amssymb,tikz-cd}
\usepackage{overpic}
\usepackage{wrapfig}
\usepackage{multirow}
\usepackage[normalem]{ulem}

\usepackage{xcolor}
\usepackage{enumitem}
\usepackage[tableposition=top]{caption}

\usepackage{nccmath}
\usepackage{caption}
\usepackage{microtype}
\usepackage{xcolor}
\usepackage{enumitem}
\usepackage[tableposition=top]{caption}
\usepackage[accepted]{icml2021}
\usepackage{multicol}
\icmltitlerunning{Learning disentangled representations via product manifold projection}
\renewcommand{\sout}[1]{} 
\colorlet{rred}{red!99}

\colorlet{red}{black}
\newcommand\mydots{\hbox to 1em{.\hss.\hss.}}

\usepackage{xcolor}
\usepackage{enumitem}

\definecolor{mygreen}{RGB}{0, 180, 0}
\definecolor{myyellow}{RGB}{250, 208, 0}
\DeclareCaptionLabelFormat{andtable}{\hspace{-0.8mm} \tablename~\thetable\,  \&  #1~#2}
\DeclareMathOperator*{\argmax}{arg\,max}
\DeclareMathOperator*{\argmin}{arg\,min}

\begin{document}

\newtheorem{defn}{\textbf{Definition}}
\newtheorem{thm}{\textbf{Theorem}}

\twocolumn[
\icmltitle{Learning disentangled representations via product manifold projection}



\icmlsetsymbol{equal}{*}

\begin{icmlauthorlist}
\icmlauthor{Marco Fumero}{to}
\icmlauthor{Luca Cosmo}{to,goo}
\icmlauthor{Simone Melzi}{to}
\icmlauthor{Emanuele Rodolà}{to}
%\icmlauthor{Fiuea Rrrr}{to}
\end{icmlauthorlist}

\icmlaffiliation{to}{Sapienza, University of Rome, Rome, Italy}
\icmlaffiliation{goo}{Università della Svizzera italiana, Lugano, Switzerland}

\icmlcorrespondingauthor{Marco Fumero}{fumero@di.uniroma1.it}

\icmlkeywords{Machine Learning, ICML}

\vskip 0.3in
]



\printAffiliationsAndNotice{}  % leave blank if no need to mention equal contribution

\begin{abstract}
% We propose a novel approach to disentangle the generator factors of a given set of data.
% We built our method on the preservation of the metric in the latent spaces that gives rise to a new definition of disentanglement.  
% On this property, we build a weakly-supervised algorithm. Indeed, as training data, we only require pairs that differ for a transformation that preserves all the latent factor except one.
% We do not require knowledge of the nature of the transformations or assumption on the properties of subspaces. We consider the standard manifold hypothesis enforced by the requirement that the manifold on which the data lives is a product of Euclidean spaces.
% We prove that our approach is easy to implement and applies to various data (2D or 3D, RGB or grayscale) and transformations of different natures.
% For these reasons, our algorithm can be injected as a plug-in module in existing models.
% We show that our approach achieves successful quantitative and qualitative results, on standard synthetic datasets and real data involved in real applications compared to state-of-the-art methods.

%
We propose a novel approach to disentangle the generative factors of variation underlying a given set of observations.
Our method builds upon the idea that the (unknown) low-dimensional manifold underlying the data space can be explicitly modeled as a product of submanifolds.
This definition of disentanglement gives rise to a novel weakly-supervised algorithm for recovering the unknown explanatory factors behind the data. At training time, our algorithm only requires pairs of non i.i.d. data samples whose elements share at least one, possibly multidimensional,  generative factor of variation.
We require no knowledge on the nature of these transformations, and do not make any limiting assumption on the properties of each subspace.
Our approach is easy to implement, and can be successfully applied to different kinds of data (from images to 3D surfaces) undergoing arbitrary transformations. In addition to standard synthetic benchmarks, we showcase our method in challenging real-world applications, where we compare favorably with the state of the art.

\end{abstract}

\section{Introduction}
Human intelligence is often understood as a process of turning experience into new behavior, knowledge, and skills \cite{Locke1689}. Humans achieve this by constructing small models of the world in their brains to explain the sensory experience and use them to infer new consequences. These models can be understood as turning experience into compact representations: functions of the experienced data which are useful for a given task.
 Representation learning \cite{bengio2014} aims to mimic this process with machines, via studying and formalizing what makes a good representation of high dimensional data, and how can we compute it in the form of an algorithm.
A central topic in this research field is  \emph{disentangled} representations, which advocates that a representation of an entity should capture the different latent factors of variation in the world where that entity is observed.
This abstract concept has led to several formalizations, but the community has not yet settled on a common definition. 

In this work, we propose a \textcolor{red}{new interpretation of disentanglement}, based on the concept of projection over product spaces, and we provide an algorithm to compute disentangled representations by a weakly supervised approach.
Since we do not require much knowledge of the data and their variations, our method fits different applications.
Our main contributions can be summarized as follows:
\begin{itemize}[topsep=0pt]
\setlength\itemsep{0.25em}
    \item We \textcolor{red}{reinterpret the notion of disentanglement in terms of }\sout{introduce a new definition of disentanglement, which relies on} geometric notions, and we show that our theoretical framework entails a generalization of current approaches.
    \item Relying on assumptions of the sample distribution, we provide a weakly supervised recipe  that is simple  to  integrate into  standard neural network models.
    \item We widely test our approach on synthetic datasets and more challenging real-world scenarios, outperforming the state of the art in several cases.
\end{itemize}

% ------------------------
\subsection{Related work}
The task of identifying the underlying factors of variation in the data has been a primary goal in representation learning research \cite{bengio2014}.

The problem of recovering independent components
in a generative process relates to the independent component analysis (ICA). Given a multivariate signal, ICA aims to separate it into additive elements generated by independent non-Gaussian sources \cite{COMON94}. While in the nonlinear case, observing i.i.d. samples, ICA was proven to be unidentifiable \cite{hyvarien1999}, recent approaches \cite{hyvarinen:2019,articleHyvarien:2016} have built on the assumption of having some degree of supervision to couple ICA and disentanglement.

A recent and promising line of research \cite{Higgins2017,kumar2018,chen:2019,kim2019} aimed to \textcolor{red}{attain disentanglement} \sout{do this} in a fully unsupervised way, relying on a probabilistic interpretation \sout{of \emph{disentanglement}} according to which a representation is defined to be  disentangled if the
data distribution can be modeled as a nonlinear transformation of a product of independent probability
distributions. Unfortunately, similarly to the ICA case, this leads to pessimistic results \cite{Locatello2019} in terms of the identifiability of the factors without any assumptions on the model or the data (inductive bias).
In this work we rephrase disentanglement in a new definition, entirely based on the geometric notion of metric and manifolds, and we rely on the assumption of observing non i.i.d. samples as data -- thus escaping the non-identifiability result of \cite{Locatello2019}.

A more general  definition of disentanglement in terms of group theory was given in \cite{Higgins2018TowardsAD}. Under their definition, a representation is disentangled if \sout{ it  matches the product of groups that define the symmetries in the world, i.e.
if the representation}it decomposes in such a way that the action of a single subgroup leaves all
factors of the representation invariant except for one.
A first practical tentative in this direction was proposed by \citet{pfau2020}, which aims to discover a decomposition of a data manifold by investigating its holonomy group; but in order to work properly, the method requires to have full access to the metric of the data manifold.
Differently, we provide a simple algorithm to obtain a disentangled representation in the form of a plug-in module for any autoencoder model.

By allowing supervision, one can escape to some extent the limitations underlined by \citet{Locatello2019}, trading off identifiability of the model with scalability to realistic settings. The latter is due to the absence of labelled data in real scenarios.
Supervision in disentanglement may come in different flavors, from weakly supervised settings \cite{locatello2020,shu2020}, to semi-supervised  \cite{locatello2020labels} and fully supervised approaches \cite{Kazemi2019,kulkarni2015deep,dubrovina2019composite, Cosmo_2020}.
We position our work in the weakly supervised setting; we assume to observe non i.i.d. sample pairs which differ only by a projection on a single submanifold.

Closer to our framework is the line of research on manifold learning, a generalization of the classical field of linear dimensionality reduction %, which generalizes factorization algorithms like PCA 
 to the nonlinear setting \cite{Tenenbaum2000,vandermaaten08,Belkin2002,Roweis2000,McInnes2018}. 
This line of works arises from the assumption that we observe the data in a high-dimensional Euclidean space $\mathcal{X}\subset \mathbb{R}^N$ sampled from a low-dimensional manifold $\mathcal{M}$, which can be approximately embedded into a low-dimensional Euclidean space $\mathcal{Z}$.
We make this assumption stronger, by conjecturing that the data manifold is factorizable into a product of subspaces; each subspace models a generative factor of the data. \textcolor{red}{ A similar assumption was adopted in \cite{shukla2019product}, where the underlying manifold is assumed to be a product of orthogonal spheres.}

\textcolor{red}{
Notably, in \cite{hu2018disentangling} it has been proposed to disentangle the factors of variation by mixing the latent vectors entries and imposing consistency constraints on the reconstructed images.
In our work we adopt a similar constraint, but we require consistency in the latent space rather than on the reconstructions. Moreover, this does not represent the main ingredient to attain disentanglement in our framework.
Finally, our work differs from all previous approaches as we consider factors of variation living on multi-dimensional submanifolds.}

%This line of works arises on the common assumption that we observe data in a high-dimensional Euclidean space $\mathcal{X}\subset \mathbb{R}^N$ is sampled from a low dimensional manifold $\mathcal{M}$, which can be approximated by embedding it in a low dimensional euclidean space $\mathcal{Z}$.
%This is typically done by constructing a neighbour graph on the sampled data in the high dimensional space and constraining the embedding in $\mathcal{Z}$ to preserve the topology of the graph, e.g. via minimizing the edge length distortion in $\mathcal{Z}$.

\section{Theoretical justification}
At the basis of our work is the assumption that high-dimensional data lie near a low-dimensional manifold, and that this can be factored into a product of submanifolds, each modelling one factor of variation. This factorization is not unique. We aim to find the one that best explains the data while being semantically valid, i.e. interpretable by humans.
We denote by $\mathcal{M} = \mathcal{M}_1 \times \mathcal{M}_2 \times ... \times \mathcal{M}_k$ the product manifold we want to approximate. In our context, we only have access to observations of $\mathcal{M} \subset \mathbb{R}^d$ in a high-dimensional space $\mathcal{X} = \mathbb{R}^N$, with $N \gg d$.

\subsection{Background}
Our definition of disentanglement is based upon the notion of metric. The manifold $\mathcal{M}$ is equipped with a metric $\mathfrak{g}_x : \mathcal{T}_x \mathcal{M} \times \mathcal{T}_x \mathcal{M} \to \mathbb{R}$, where $\mathcal{T}_x\mathcal{M}$ is the tangent space at $x\in\mathcal{M}$; $\mathfrak{g}_x$ is a smoothly varying function that enables the computation of lengths, angles and areas on the manifold.
In particular, the distance between two arbitrary points $x,y$ on the manifold is defined in terms of the metric tensor: $d^2(x,y) =\int_0^1 \mathfrak{g}_{\gamma_t}(\dot{\gamma}_t,\dot{\gamma}_t )dt$ where $\gamma :[0,1] \to \mathcal{M}$ is a geodesic parametrized curve with $\gamma(0)=x$ and $\gamma(1)=y$.

Since the metric allows us to compute distances, we can alternatively regard $\mathcal{M}$ as a metric space.
To define distances on a {\em product} manifold, we recall the following definition for product metric spaces:

\begin{defn}[Product metric space]
A product metric space is an ordered pair $(M,d)$ where $M$ is a product of sets $M_1 \times ... \times M_k$, each equipped with a function  $d_i:M_i \times M_i \to \mathbb{R}, \ \forall i$ s.t. $\forall x,y,z \in M_i$:
\begin{align*}
    & d_i(x,y) = 0 \iff x = y \\
    & d_i(x,y) = d_i(y,x)\\
    & d_i(x,z) \leq d_i(x,y) + d_i(y,z)
\end{align*}
The metric on the product space corresponds to the $L_2$ norm of the metric on the subspaces.
\end{defn}

% A set $M$ is a metric space if it is equipped with a symmetric function  $d:M \times M \mapsto \mathbb{R}$ for which satisfies the identity of indiscernibles condition and the triangle inequality.

Distances on a product manifold can then be factored into the sum of distances on the submanifolds, since the metric of the product manifold simply reduces to the sum of the metrics of its constituents $\mathfrak{g}(u,v) = \sum_i^K \mathfrak{g}_i(u_i,v_i)$, resulting in a product metric space \cite{Lee2012}.

\begin{figure}[t]
    \centering
    \begin{overpic}[trim=0cm 0.5cm 1cm 0.5cm,clip,width=\linewidth]{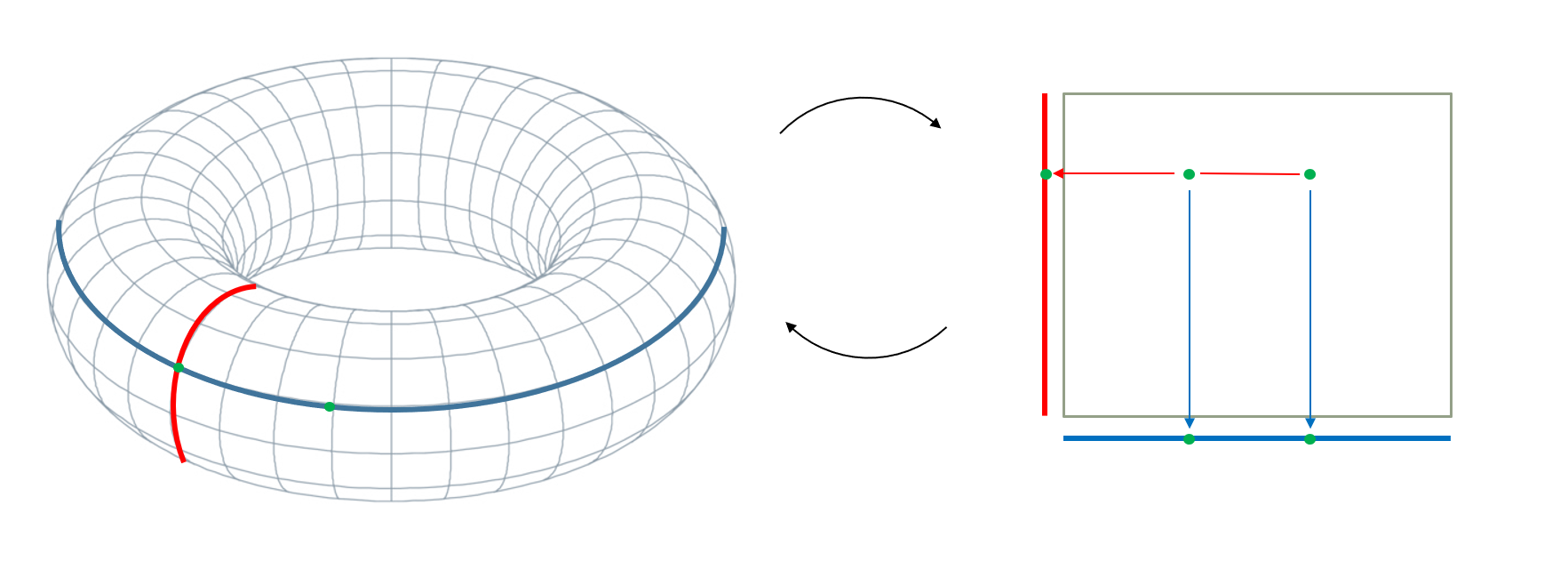}
    \put(10.1, 30){\footnotesize $\mathcal{M}$ }
    \put(-1, 21){\color{blue}\footnotesize $\mathcal{M}_2$ }
    \put(13, 12){\color{rred}\footnotesize $\mathcal{M}_1$ }
    \put(20, 6.5){\scriptsize $x_2$ }
    \put(63, 18){\color{rred}\footnotesize $\mathcal{S}_1$ }
    \put(82, 2){\color{blue}\footnotesize $\mathcal{S}_2$ }
    \put(93, 30){\color{black}\footnotesize $\mathcal{Z}$ }
    
    \put(79, 26){\scriptsize $z_2$ }
    \put(88, 26){\scriptsize $z_1$ }
    
    %\put(65, 26){\color{red}\scriptsize $\Pi_{\mathcal{S}_1} (z_1)$ }

    \put(79, 26){\scriptsize $z_2$ }
    \put(88, 26){\scriptsize $z_1$ }
    
    \put(20, 6.5){\scriptsize $x_2$ }
    \put(11, 8){\scriptsize $x_1$ }
    \put(56, 30){\footnotesize $\tilde{f}$ }
    \put(56, 8){\footnotesize $\tilde{f}^{-1}$ }
    \end{overpic}
    \caption{Let $\mathcal{X} \subset \mathbb{R}^3$ be our input space composed by points lying near a 2-dimensional torus $\mathcal{M}$. Since the torus is a product manifold of two circles, we can represent points on the torus as projections on the two submanifolds $\mathcal{M}_1$ and $\mathcal{M}_2$.  Let $\tilde{f}:\mathcal{X} \to {Z}$  be an embedding into a subset of the plane $\mathcal{Z} \subset \mathbb{R}^2$. $\mathcal{Z}$ can be factored into two subspaces $\mathcal{S}_1  \subset \mathbb{R}$ and $\mathcal{S}_2  \subset \mathbb{R}$, which reflect the metric structure of $\mathcal{M}_1, \mathcal{M}_2$.}
    \label{fig:torus}
\end{figure}
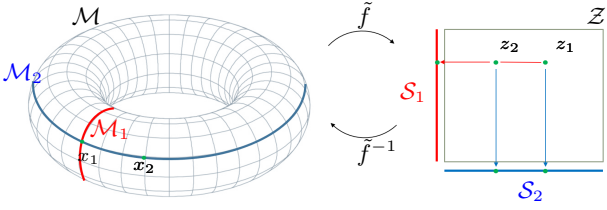

\subsection{Disentangled representations}

We define a representation as \emph{disentangled} if the variation of one generative factor in the data corresponds to a change in {\em exactly one} submanifold. 
Let us be given two data samples, and assume one of them is transformed by some unknown process. Then, the change in the distance between their projections onto the product manifold, measured before and after transforming the sample, is nonzero only on one submanifold. 
In other words, changing one factor of variation at a time will correspond to moving along a trajectory on a specific submanifold, while standing still on all the others.
More formally:
\begin{defn}
 Let $\mathcal{M} = \mathcal{M}_1 \times \mathcal{M}_2 \times \ldots \times \mathcal{M}_k$ be a product manifold embedded in high dimensional space $\mathcal{X}$. A representation $z$ in some space $\mathcal{Z}=S_1 \times \ldots \times S_k$, \textcolor{red}{s.t. $dim(\mathcal{Z}) \ll dim(\mathcal{X})$,} is {\em disentangled} w.r.t. $\mathcal{M}$ if there exists a smooth invertible function, with smooth inverse (a diffeomorphism)  $\tilde{g}:\mathcal{Z} \to \mathcal{M}$ %if there exists an embedding map $\tilde{f}$ between $\mathcal{X}$and $\mathcal{Z}$ 
 s.t. $\forall i \in {1, \ldots , k};  \ \forall  x_1, x_2 \in \mathcal{M} $:
 \begin{align*}
    & d_{\mathcal{M}_i}(x_1^i,x_2^i) >0 \implies d_{\mathcal{S}_i}(s_1^i,s_2^i) >0 \\
    &  d_{\mathcal{M}_i}(x_1^i,x_2^i) = 0 \implies d_{\mathcal{S}_i}(s_1^i,s_2^i) = 0 \implies s_1^i = s_2^i 
\end{align*}
where $x_j^i$ is the projection of $x_j$ on $\mathcal{M}_i$ and $s_j^i= \Pi_i \tilde{g}^{-1}(x_j)$ with $\Pi_i$ being the projection onto the subspace $\mathcal{S}_i \subset \mathcal{Z}$.
\label{def:2}
\end{defn}
The existence of $\tilde{g}^{-1}$ is guaranteed by the Whitney immersion theorem \textcolor{red}{\cite{Whitney:36}}, which states that for any manifold of dimension $m$ there exists a one-to-one mapping in a Euclidean space of dimension at least $2m$ ($2m-1$ if the manifold is smooth). Note that, if one has access to the metric of $\mathcal{M}_i$, a stricter definition of disentanglement can be enforced, requiring that the metric on the subspaces fully recovers the one on the submanifolds, i.e. $          d_{\mathcal{M}_i}(x_1^i,x_2^i) \approx d_{\mathcal{S}_i}(s_1^i,s_2^i) $. 
 The Nash-Kuiper embedding theorem \textcolor{red}{\cite{Nash:56}} ensures the existence of the isometry $\tilde{g}^{-1}$ in the latter case. Unfortunately, knowing the metric of $\mathcal{M}_i$ is unlikely in a practical setting. %this can be done isometrically, i.e. without any metric distortion, which addresses the strongest version of the definition.

\textbf{Example.}
To better understand the definition we refer the reader to Figure~\ref{fig:torus}, where we also introduce some notation.
Let $\mathcal{X} \supset \mathcal{M}$ be our observation space, equipped with the standard Euclidean metric $d_e(\cdot,\cdot)= ||\cdot - \cdot ||_2$. We consider the embedding function $\tilde{f}:\mathcal{X} \to \mathcal{Z}$, \textcolor{red}{s.t. $\tilde{f} \approx \tilde{g}^{-1}$ on $\mathcal{M}$}, where $\mathcal{Z}$ is equipped with the metric induced via $\tilde{f}$ and $\mathrm{dim}(\mathcal{Z}) \ll \mathrm{dim}(\mathcal{X})$.
We consider the factorization $\mathcal{Z} = \mathcal{S}_1 \times ... \times \mathcal{S}_K$, where each subspace ideally corresponds to a parametric space for each of the $\mathcal{M}_i \subset \mathcal{M}$.
We name the projections onto $\mathcal{S}_i$ as $\Pi_i: \mathcal{Z} \to \mathcal{S}_i  \ \forall i \in \left\{1...K \right\}$. 
In Figure~\ref{fig:torus}, the  subsets $\mathcal{S}_1,\mathcal{S}_2$ approximate the metric structure of $\mathcal{M}_1,\mathcal{M}_2$ in the sense of Definition~\ref{def:2}: $z_1, z_2$ projected onto $\mathcal{S}_1$ will correspond to the same point (i.e. their distance will be zero), reflecting the structure of $\mathcal{M}_1$, while their projection onto $\mathcal{S}_2$ will approximate $d_{\mathcal{M}_2}(x_1,x_2)$.%the distance of $x_1, x_2 $ on $\mathcal{M}_2$.

Furthermore, we can regard $x_2$ as the result of a transformation $T$ applied to $x_1$, namely a translation on the submanifold $\mathcal{M}_2$ (in general, each pair of samples from $\mathcal{M}$ can be seen in this sense). This allows us to describe the example of Figure \ref{fig:torus} via the commutative diagrams in Figure \ref{fig:comm_diagrams}.

\begin{figure}[!h]
   \begin{minipage}{0.48\linewidth}
    \centering
    \begin{tikzcd}
x_1 \arrow[r, "\tilde{f}"] \arrow[d, "T"'] & z_1 \arrow[d, "T_z"] \arrow[r,"\Pi_1"] & s_1^1 \arrow[d, "Id"']  \\
x_2 \arrow[r, "\tilde{f}"']                & z_2       \arrow[r,"\Pi_1"]          & s_1^2
\end{tikzcd}
   \end{minipage}%
   \begin{minipage}{0.48\linewidth}
    \centering
    \begin{tikzcd}
x_1 \arrow[r, "\tilde{f}"] \arrow[d, "T"'] & z_1 \arrow[d, "T_z"] \arrow[r,"\Pi_2"] & s_2^1 \arrow[d, "T_{s_2}"']  \\
x_2 \arrow[r, "\tilde{f}"']                & z_2       \arrow[r,"\Pi_2"]          & s_2^2
\end{tikzcd}
   \end{minipage}
    \caption{Commutative diagrams describing the relation between the transformation $T$ on $\mathcal{M}$ that moves $x_1$ to $x_2$, the corresponding transformation $T_{z}$ on $\mathcal{Z}$, and after the projections $\Pi_1$ (left diagram) and $\Pi_2$ (right diagram) on the subspaces $\mathcal{S}_1$ and $\mathcal{S}_2$ respectively.}
    \label{fig:comm_diagrams}
\end{figure}

\textbf{Can we learn disentangled representations?}
Our main objective is to obtain a disentangled representantion (with respect to Definition \ref{def:2}) without having any access to neither the metrics, nor the manifold $\mathcal{M}=\mathcal{M}_1 \times \ldots \times\mathcal{M}_k $ or its subspaces. We only assume to observe pairs of samples $(x_1,x_2)$ in the high-dimensional data space $\mathcal{X}$. 

%Our assumption is that one of the two elements composing the pair results from a transformation of the other, $x_2 = T_i(x_1)$, which corresponds to a translation over the submanifold $\mathcal{M}_i$, which is  invariant to transformations $T_j$ over $\mathcal{M}_j\forall j \neq i$ (i.e. the distance of the projection of the sampled pair on the corresponding submanifolds $\mathcal{M}_j$ is zero $\forall j$).
Our assumption is that w.l.o.g. $x_2$ results from a transformation of $x_1$, $x_2 = T_i(x_1)$, corresponding to a translation over the submanifold $\mathcal{M}_i$ for some $i$. Moreover, \sout{we require that $(x_1,x_2)$ is invariant to transformations $T_j$ over $\mathcal{M}_j \ \forall j \neq i$,} \textcolor{red}{the projection of $(x_1,x_2)$ on the submanifold $\mathcal{M}_i$ is invariant to transformations $T_j$ over $\mathcal{M}_j$}. Specifically, the distance of the projection of the sampled pair on the corresponding submanifolds $\mathcal{M}_j$ is zero $\forall j \neq i$.\footnote{\textcolor{red}{In practice, we relax this assumption to the case where multiple transformations can occur at the same time, i.e. $x_2=T^*(x_1)$, where $T^* = T_1 \circ ...  \circ T_{k}$ and at least one $T_i =Id,  \ i=1 \ldots k$.}}

Then, our objective is to learn the mapping $\tilde{f}:\mathcal{X} \to \mathcal{Z}$, s.t. $\tilde{f} \approx g^{-1}$ on $\mathcal{M}$, where $\mathcal{Z}$ is a low-dimensional product space which acts as a parametric space for $\mathcal{M}$. Therefore, the mapping $\tilde{f}$ composed with the projections $\Pi_i \ \forall i \in {1, \dots,k}$ must be \emph{invariant} with respect to transformations $T_j$, and \emph{equivariant} with respect to the transformation $T_i$.
We answer the titular question in the following.

\textbf{Relation with \cite{locatello2020}.}
The theoretical setting of  \citet{locatello2020} is a special case of our framework, where their $\mathcal{S}_i$ are only $1$-dimensional, the $s \in S_i$ parametrize a fixed (tipically Gaussian) distribution $P(s_i)$, and $d(a,b)= KL(P(s') \| P(s)) \  \forall s,s' \in \mathcal{S}$ is the Kullback-Leibler divergence.
%The theoretical setting  in the work of  \cite{locatello2020} can be considered as a special case of our theoretical framework, where the $\mathcal{S}_i$ are 1 dimensional, their $s \in S_i$ parametrize a fixed  one dimensional (tipically gaussian) distribution $P(s_i)$ and $d(a,b)= KL(P(s') || P(s)) \  \forall s,s' \in \mathcal{S}$, where $KL$ is the Kullback-Leibler divergence. 
Assuming to have a sufficiently dense sampling of the data, and given that the map $\tilde{g}$ in Definition~\ref{def:2} is a diffeomorphism, these conditions are sufficient to prove identifiability for our model, reconnecting to the exact case proved in \citet{locatello2020}.  
%se l'assunzione di diffeomorfismo locale per g regge possiamo ricondurci a lcatello e provare identificabilità.

%\textbf{Relation with \cite{Higgins2018TowardsAD}}

% \begin{figure}[h!]
%     \centering
%     \begin{tikzcd}
% x_1 \arrow[r, "f"] \arrow[d, "T"'] & z_1 \arrow[d, "T_z"] & \\
% x_2 \arrow[r, "f"']                & z_2                 
% \end{tikzcd}
%     \caption{Commutative diagram 1}
%     \label{fig:my_label}
% \end{figure}

\begin{figure}[t!]
\centering
    \begin{overpic}[trim=0cm 0.2cm 0cm 0.22cm,clip,width=0.95\linewidth]{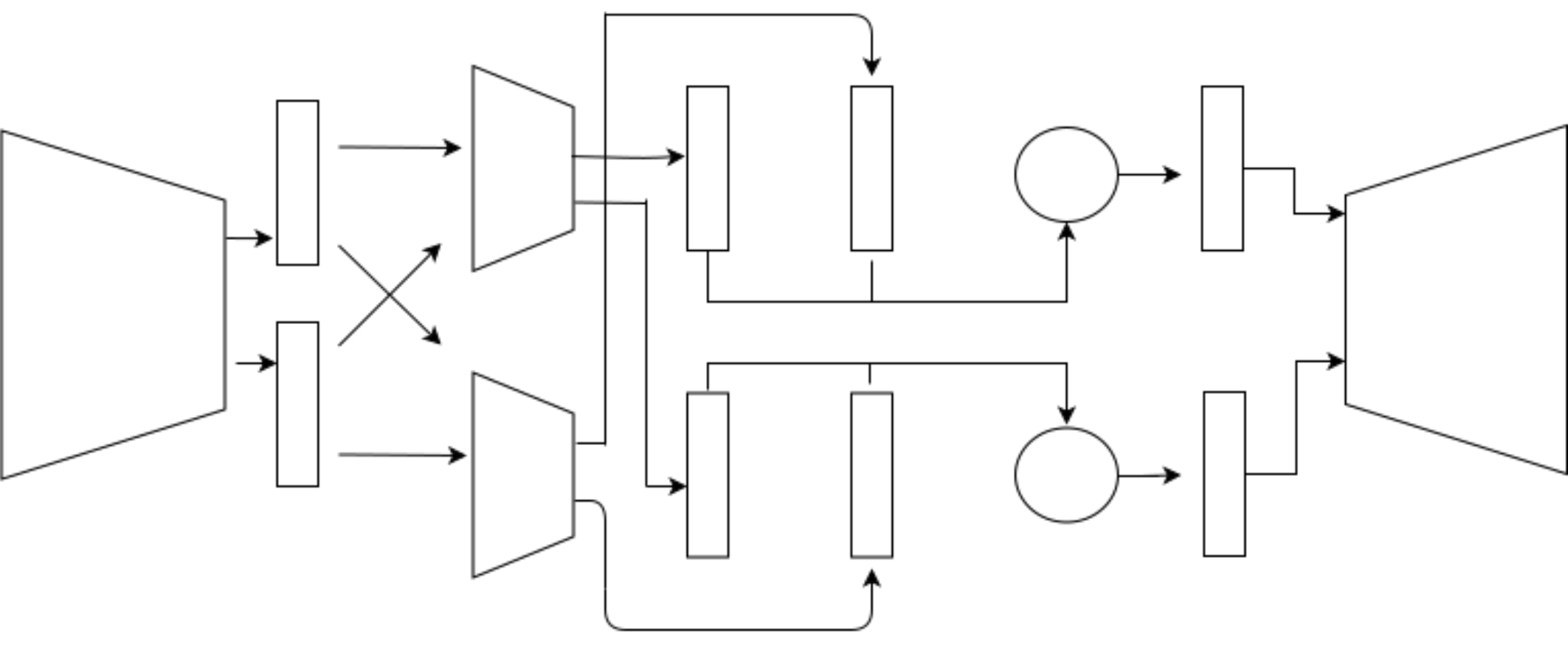}
     
    \put(-4.4, 29.5){\footnotesize $x_{1}$ }
    \put(-4.4, 13.5){\footnotesize $x_{2}$ }
    \put(31, 29.7){\footnotesize $P_{1}$ }
    \put(33, 20){$\cdot$ }
    \put(33, 19){$\cdot$ }
    \put(33, 18){$\cdot$ }

    \put(31, 10){\footnotesize $P_{k}$ }
    \put(6, 21.5){\footnotesize $f$ }
    \put(92, 21.5){\footnotesize $g$ }
    \put(17.4, 36.5){\footnotesize $\hat{z}_1$ }
    \put(17.4, 6){\footnotesize $\hat{z}_2$ }
    \put(44.2, 37){\footnotesize $s_1^1$ }
    \put(49, 30){$\cdot$ }
    \put(50, 30){$\cdot$ }
    \put(51, 30){$\cdot$ }
    \put(56.5, 37){\footnotesize $s_1^k$ }
    \put(44.2, 1.5){\footnotesize $s_2^1$ }
    \put(49, 8.5){$\cdot$ }
    \put(50, 8.5){$\cdot$ }
    \put(51, 8.5){$\cdot$ }
    \put(56.5, 1.5){\footnotesize $s_2^k$ }
    \put(76.4, 36.5){\footnotesize $z_1$ }
    \put(76.4, 1.5
    ){\footnotesize $z_2$ }
     
    \put(102, 29.5){\footnotesize $\hat{x}_{1}$ }
    \put(102, 13.5){\footnotesize $\hat{x}_{2}$ }
    \put(66.0, 29){\tiny aggr}
    \put(66.0, 9.8){\tiny aggr}
     
   \end{overpic}
\caption{
The architecture of our model. We process data in pairs $(x_1,x_2)$, which are embedded into a lower dimensional space $\mathcal{Z}$ via a twin network $f$. The image $(z_1,z_2)$ is then mapped into $k$ smaller spaces $\mathcal{S}_1, \dots ,\mathcal{S}_k \subset\mathcal{Z}$ via the nonlinear operators $P_i$. The resulting vectors are aggregated in $\mathcal{Z}$, \textcolor{red}{with $aggr= +$}, and mapped back to the input data space by the decoder $g$, \textcolor{red}{to get $(\hat{x_1},\hat{x_2})$}. As we do not impose any constraint on $f$ and $g$, the intermediate module of the architecture could be attached to any autoencoder model. \textcolor{red}{For a zoomed-in version see the Appendix.}
% The architecture of our model. We process data in pairs $(x_1,x_2)$ which are embedded in lower dimensional space $\mathcal{Z}$ via a siamese network $f$. The image $(z_1,z_2)$ is then projected into $k$ subspaces $S_1,..S_k$ via the nonlinear projections $P_i$. Projected vectors are aggregrated and mapped back to the input space by the decoder $g$. As we do not put any constraint on the choice of $f$ and $g$, the intermediate module can be in pricipled attached to any autoencoder model.
\label{fig:architecture}}
\end{figure}
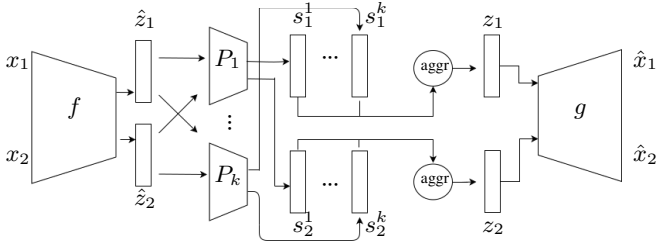

\section{Method}
%% HOLD %%
% The main question we try to answer in this paper is if a disentangled representantion, with respect to definition \ref{def:2} can be obtained without having any access to the metrics nor the manifold $\mathcal{M}=\mathcal{M}_1 \times \ldots \times\mathcal{M}_k $ or its subspaces themselves, but only by observing pairs of samples $(x_1,x_2)$ in the high dimensional space $\mathcal{X}$. 

To implement our framework in practice, we approximate $\tilde{f}$ and its inverse with an encoder-decoder model where the encoder $f \approx \tilde{g}^{-1} $, and the decoder $g \approx \tilde{g}$. 
\textcolor{red}{We impose the product space structure $\mathcal{Z}=\mathcal{S}_1 \times \ldots \times \mathcal{S}_k$ on $\mathcal{Z}$ by adding a projection module in the latent space.} This module is composed of nonlinear operators $P_i \ \forall i \in {1,\dots,k}$, one for each factor of variation (we discuss the choice of $k$ in the experimental section). The $P_i$ act as nonlinear projectors, mapping from an  \emph{entangled}, intermediate latent space $\hat{\mathcal{Z}}$ onto the  \emph{disentangled} %, and are the composition of a nonlinear transformation from $Z$ to a space $\hat{z}$, and a linear projection $\Pi_i$ onto the
 subspaces $\mathcal{S}_i \subset{\mathcal{Z}}$. We remark that the  subspaces $\mathcal{S}_i$ act as parametric spaces for the latent submanifolds $\mathcal{M}_i \subset \mathcal{M}$, where each $\mathcal{M}_i$  characterizes a single factor of variation, in the sense of Definition~\ref{def:2}. \sout{The $P_i$ act as nonlinear projectors onto the %, and are the composition of a nonlinear transformation from $Z$ to a space $z$, and a linear projection $\Pi_i$ onto the
 subspaces $\mathcal{S}_i$. }%, and correspond to the composition $\Pi_i \circ f$ described in the theoretical part, see Figure~\ref{fig:comm_diagrams}. %This operators, together with additional terms to be added to the loss function, described in the following subsection. 
 
Our proposed architecture is illustrated  in Figure~\ref{fig:architecture}.
%In the following, we describe in detail our architecture, 
%forse mettere qua le proprietà che vogliamo che queste spazio fattorizzato abbia, e poi spiegare perchè le loss le impongono
%  illustrated  in Figure~\ref{fig:architecture}.
% \textcolor{red}{We choose to impose  the structure of a Cartesian product space to the subspaces $\mathcal{S}_1 \times \ldots \times \mathcal{S}_k= \mathcal{Z}$,  coupling a sparsity constraint imposed by a specific term in our loss function, together with the choice of  the sum as the aggregation operator for the subspaces (other choices for $aggr$ may be possible according to the desired structure for $\mathcal{Z}$).}
\textcolor{red}{As we model $\mathcal{Z}=\mathcal{S}_1 \times \ldots \times \mathcal{S}_k$ as a product vector space, a latent vector in $\mathcal{Z}$ is obtained by aggregating the subspaces through a concatenation operation. Note that, in principle, the subspaces $\mathcal{S}_i$ could have different dimensionalities, which we are interested in learning. In practice, we set the same dimensionalities $dim(\mathcal{S}_i)=dim(\mathcal{Z})=dim(\hat{\mathcal{Z}})$ for each subspace  $\mathcal{S}_i$ and add a sparsity and orthogonality constraint to them, enforced by a specific term in our loss function. This results in the projectors $\mathcal{P}_i$ mapping onto disjoint subsets of the dimension indices for each subspace, thus allowing us to approximate the concatenation operation through the sum of the subspaces (proof in Appendix), which corresponds to set the operator $\operatorname{aggr}=+$.  We remark that other choices for $aggr$ may be possible according to the desired structure to impose on $\mathcal{Z}$.}% In the following we describe individually the terms composing our loss function, detailing how each of them contributes to obtain a disentangled representation of the data.}
% To promote the 

\subsection{Losses}
% We consider $f$ be parametrized by a set of parameters $\theta$ and  $g$ be parametrized by $\gamma$.
We model $f$ and $g$ as neural networks parametrized by a set of parameters $\theta$ and $\gamma$ respectively, while the nonlinear operators 
% We approximate projections $\Pi_i $ with nonlinear operators $P_i, \forall i$, parametrized by $\omega$.
%Instead of explicitly expressing $P_i$ as compositions $\Pi_i \circ f$, we model them as networks 
 $P_i$ are parametrized by \sout{$\omega$} \textcolor{red}{$\omega_i$}. 
The model is trained by minimizing the composite energy $\mathcal{L}$: 
\begin{align}
    \mathcal{L} = \mathcal{L}_{rec} + \beta_1 (\mathcal{L}_{dis}+\mathcal{L}_{spar}) +\beta_2 \mathcal{L}_{cons} + \beta_3\mathcal{L}_{reg}
\end{align}
\textcolor{red}{w.r.t. the parameters $\theta,\gamma,\omega_i$}, balanced by regularization parameters $\beta_1,\beta_2,\beta_3$. 
% We're going to explain them in more detail in the following subsections.
We now provide an explicit formula for each term and describe their role in the optimization.

\textbf{Reconstruction loss}
\textcolor{red}{
\begin{align}
         & \mathcal{L}_{rec}\hspace{-0.1cm}=\hspace{-0.1cm}\|x - g_\gamma(\operatorname{aggr}( P_{1,\omega_1}f_\theta(x), \mydots, P_{k,\omega_k}f_\theta(x)))\|_2^2  
         %& \mathcal{L}_{rec}=||x - g_\gamma(\operatorname{aggr}_{i=1}^k( P_{i,\omega_i}f_\theta(x)))||_2^2 
\end{align}}
\sout{ old
\begin{align}
         & \mathcal{L}_{rec}=\argmin_{\theta,\gamma}\mathbb{E}_x ||x - g_\gamma(f_\theta(x))||_2^2 
\end{align}}The reconstruction term captures the global structure of the manifold $\mathcal{M}$ by enforcing $f \approx g^{-1}$. We remark that invertibility also implies bijectivity on the data manifold.

\textbf{Consistency losses}
% \begin{align}
%      &\mathcal{L}_{cons}(s_i)=\argmin_{\theta,\omega_i}||P_{i,\omega_i}(f_\theta(x_{s_i})) - s_i||_2^2, \\  
%     &\text{where } x_{s_i} = g(aggr( P_i(x_1),...,P_j(x_2))), \ \forall i \in {1, \ldots , k} \notag
% \end{align}

\begin{align}
     &\mathcal{L}_{cons}=\sum_{i=1}^{k}||P_{i,\omega_i}(f_\theta(\hat{x}_{s_i})) - s_i||_2^2, 
    %  \\  
    % &\text{where } x_{s_i} = g(aggr( P_i(x_1),...,P_j(x_2))), \ \forall i \in {1, \ldots , k} \notag
    \label{eq:consistency}
\end{align}
with \textcolor{red}{$s_i\hspace{-0.08cm}=\hspace{-0.08cm}P_if(x_1)$  and}
\textcolor{red}{$ \hat{x}_{s_i} \hspace{-0.15cm}=\hspace{-0.10cm}g (\mathrm{aggr}( P_i f(x_1),P_{j \neq i}f(x_2)))$.}%, where $P_{\mathcal{S}_{j \neq i}}:=\{ \bigoplus_{j \in \{ 1 \ldots k\}} P_j| j \neq i\}$} and $\bigoplus$ denoting the concatenation operator.

%$ \hat{x}_{s_i} \hspace{-0.15cm}=\hspace{-0.10cm}g (\mathrm{aggr}( \{P_i f(x_1)\} \cup \{ P_{j}f(x_2) | j\in {1...k}, j\neq i \} ))$

%\textcolor{red}{$ \hat{x}_{s_i} = g (\mathrm{aggr}( P_i(f(x_1)),P_l(f(x_2)),\ldots ,P_j(f(x_2)))),$ $\forall j,l \in {1, \ldots , k}$, with $l \neq i$ and $j \neq i$.}
%\sout{
%\begin{align}
%      &\mathcal{L}_{cons}=\argmin_{\theta,\{\omega_i\}_{i=1}^{k}}\sum_{i=1}^{k}||P_{i,\omega_i}(f_\theta(x_{s_i})) - s_i||_2^2, 
%     %  \\  
%     % &\text{where } x_{s_i} = g(aggr( P_i(x_1),...,P_j(x_2))), \ \forall i \in {1, \ldots , k} \notag
%     \label{eq:consistency}
% \end{align}
% where \sout{$ x_{s_i} = g(\mathrm{aggr}( P_i(x_1),P_l(x_2),...,P_j(x_2))), \ \forall j,l \in {1, \ldots , k}$, with $l \neq i$ and $j \neq i$.}

%\textcolor{red}{$ x_{s_i} = g(\mathrm{aggr}( P_i(f(x_1)),P_l(f(x_2)),...,P_j(f(x_2)))), \ \forall j,l \in {1, \ldots , k}$, with $l \neq i$ and $j \neq i$. The $aggr$ operator corresponds to the sum.}}

The minimization of this loss makes the nonlinear operator $P_i$ invariant to changes in the subspaces $S_j, \forall j \neq i $.
\setlength{\columnsep}{7pt}
\setlength{\intextsep}{1pt}
\begin{wrapfigure}[10]{r}{0.5\linewidth}
\vspace{-0.3cm}

\begin{center}
\begin{overpic}
[trim=0cm 0cm 0cm 0cm,clip,width=1.0\linewidth]{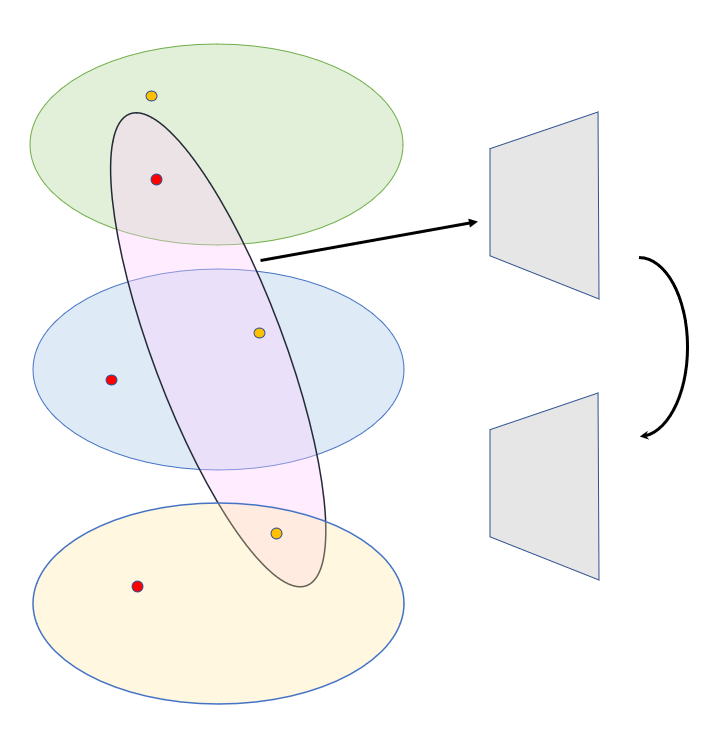}
\put(40,82){ \color{mygreen}\footnotesize $\mathcal{S}_i$}
\put(41,43){ \color{blue}\footnotesize $\mathcal{S}_j$}
\put(40,13){ \color{myyellow}\footnotesize $\mathcal{S}_l$}

\put(71,71){\footnotesize $g$}
\put(71,32){\footnotesize $f$}
\put(82,71){\footnotesize $\hat{x}_{s_i}$}
\put(22,72){\scriptsize  $P_{i}(f(x_1))$}
\put(36,50.7){\scriptsize $P_{j}(f(x_2))$}
\put(38,24.6){\scriptsize $P_{l}(f(x_2))$}
\end{overpic}
\end{center}
\end{wrapfigure}
This invariance is induced by aggregating, at training time, the representations $P_i(x_1)$ with the images of $x_2$ through $P_j, \ \forall j \neq i$ in the latent space $\mathcal{Z}$.
This combination creates a new latent vector that differs from the encoding of $x_1$ only in the subspaces $\mathcal{S}_j, \forall j \neq i$ as illustrated in the inset Figure.
Forcing \sout{$P_i(x_1)$} \textcolor{red}{$P_i(f(x_1))$} and the image of the resulting latent vector via the composition $P_i\circ f \circ g$ to coincide in $\mathcal{S}_i$, is equivalent to require that $P_i$ is invariant to changes in the other subspaces $\mathcal{S}_j$; this promotes injectivity of the $P_i$.

\textbf{Distance loss.}
This term, defined below, 
ensures the second property of Definition~\ref{def:2} and, coupled with the consistency constraint, is the key loss for constraining disentanglement in the representation.

Given a pair $(x_1,x_2)$ and its latent representation $(z_1,z_2)$, assume the existence of an oracle $\mathcal{O}$, that, acting on the latent vectors, tells us exactly which subspace $\mathcal{S}_i$ of  $\mathcal{Z}$ contains the difference in the input pair.
If $x_2 = T_i (x_1)$ for a fixed $T_i$ acting on $\mathcal{M}_i$ then $\mathcal{O}(\hat{z_1},\hat{z_2})= i$.
% Considering the transformation $T$ which generates $x_2$ from $x_1$, $S_i$ will correspond to the space on which $T$ acts, i.e. $\mathcal{O}(z1,z2)= i$ if $x_2 = T_i (x_1)$ for a fixed $T_i$ acting on $\mathcal{M}_i$.
The oracle $\mathcal{O}$ can be implemented in practice by allowing a higher degree of supervision, i.e. by incorporating labels into the sampled data pairs, as done in \cite{Zhan:2019,locatello2020labels}.
However, in real settings, typically we do {\em not} have access to labels, thus $\mathcal{O}$ has to be estimated.

To estimate $\mathcal{O}$ in our weakly-supervised setting, we proceed as follows:
(i) We compute a distance between the projected pair $d(s^i_1,s^i_2)$ in each subspace $\mathcal{S}_i$; (ii) we estimate the oracle as $\tilde{\mathcal{O}}(\hat{z_1},\hat{z_2})=\argmax_i d(P_i(\hat{z_1}),P_i(\hat{z_2}))$, that corresponds to the  subspace $\mathcal{S}_i$ where the projections differ the most.
%$\tilde{\mathcal{O}}$ is our approximation of the unknown $\mathcal{O}$.
% (i) We compute a measure of similarity between the projected pair $d(s^i_1,s^i_2)$ , in each subspace $\mathcal{S}_i$; (ii) We fix the subspace $i$ in which the projections differ the most;
% % where the least similarity is realized (i.e. the subspace 
% (iii) We estimate the unknown $\mathcal{O}$ as $\tilde{\mathcal{O}}(z_1,z_2)=\argmax_i d(P_i(z1),P_i(z_2))^2$.
%
 The distance in (i) should be general enough to be compared in the different subspaces. We select the Euclidean distance normalized by the average length of a vector in each subspace to provide a reliable measure, also considering different dimensions. We indicate with $\delta_i=d(P_{i, \omega_i}(\hat{z_1}),P_{i,\omega_i}(\hat{z_2}))$ the \sout{squared} distance between  the images on the $i$-th subspace, $\forall i \in 1, \ldots, k$.

We constrain the images onto the subspaces not selected by $\tilde{\mathcal{O}}$ to be close to each other, because they should ideally correspond to the same point as visualized in the inset Figure.
\setlength{\columnsep}{6pt}
\setlength{\intextsep}{1pt}
\begin{wrapfigure}[6]{r}{0.35\linewidth}
\vspace{-0.25cm}

\begin{center}
\begin{overpic}
[trim=0cm 0cm 0cm 0cm,clip,width=0.9\linewidth]{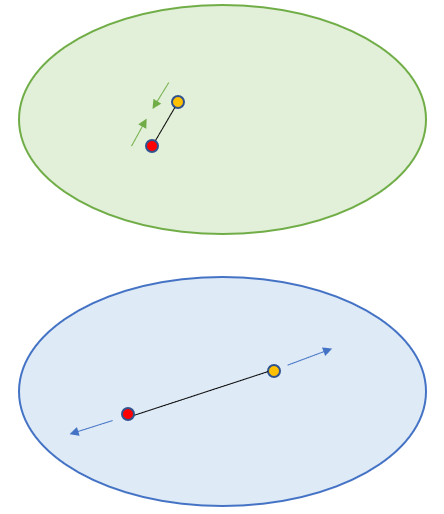}
\put(38,80){\tiny $P_j(\hat{z}_2)$}
\put(25,62){\tiny $P_j(\hat{z}_1)$}

\put(45,35){\tiny $P_i(\hat{z}_2)$}
\put(15,10){\tiny $P_i(\hat{z}_1)$}

\put(65,70){ \color{mygreen}\footnotesize $\mathcal{S}_j$}
\put(65,16){ \color{blue}\footnotesize $\mathcal{S}_i$}
\end{overpic}
\end{center}
\end{wrapfigure}
To avoid the collapse of multiple subspaces into a single point, we insert a contrastive term \cite{Hasell:2006} to balance the loss, which encourages the projected points in $\mathcal{S}_i$ to move away from each other.
The resulting distance loss is therefore written as:
% The measure of similarity should be general enough to be compared in the different subspaces. To provide a reliable measure also considering higher dimensions, we choose the squared euclidean distance normalized by the average length of a vector in each subspace. Indicating with $D_i=d(P_{i,\omega_i}(z_1),P_{i,\omega_i}(z_2))^2$ the squared distance between  the images on the $i$-th subspace, we can write the distance loss as:
%the cosine similarity:
%\begin{align*}
%    \text{cos sim}(s_i^1,s_i^2) = \frac{\langle s_i^1,s_i^2 \rangle }{||s_i^1||\cdot||s_i^2||}
%\end{align*}

\begin{equation}\label{Eq:dist_loss}
        %&\text{Let }\tilde{i} = %\argmax_i \  D_i \\
         %&\mathcal{L}_{dis}(z_1,z_2)=\\
        %&\argmin_{\theta,\omega_j}||P_{j,\omega_j}(x_1)- mean(P_j(f(x_1)),P_j(f(x_2)))||_2^2
        %&\argmin_{\theta,\omega_j}||P_{j,\omega_j}(x_1)- P_{j,\omega_j}(x_2)||_2^2  +
        \mathcal{L}_{dis}=\hspace{-0.03cm} \sum_{i=1}^{k} \hspace{-0.03cm} (1\hspace{-0.06cm}-\hspace{-0.06cm}\alpha_i) \delta_{i}^2+ \alpha_i\hspace{-0.06cm}\max(m\hspace{-0.06cm}-\hspace{-0.06cm} \delta_{i},\hspace{-0.05cm}0)^2\,,
 \end{equation}
%  where $p$ is an indicator variable which tell if the $i$-th subspace $\mathcal{S}_i$ encodes the generative factors which are not shared between the input pair. In other words, we have $p= 1 $ if $\tilde{\mathcal{O}}(z_1,z_2)=i$ and $p=0 $ otherwise.
 where $\alpha_i= 1 $ if $\tilde{\mathcal{O}}(\hat{z_1},\hat{z_2})=i$ and \sout{$p=0 $} \textcolor{red}{$\alpha_i=0$} otherwise;  $m$ is a fixed margin, which constrains the points to be at least at distance $m$ from each other, and prevents the contrastive term from being unbounded.
 
\textbf{Sparsity Loss}
\begin{align}
        &\mathcal{L}_{spar}= \sum_{i=1}^{k}\| P_{i,\omega_i}(f_{\theta}(x))\odot \sum_{j\neq i}^k P_{j,\omega_j}(f_{\theta}(x)) \|_1\,,
\end{align}
where $\odot$ denotes the element-wise product.

This loss combines an orthogonality and an $\ell_1$-sparsity constraint on the latent subspaces.
It allows us to set only the dimensionality of the latent space $\mathcal{Z}$ as an hyperparameter, while the algorithm infers the dimensionalities of the subspaces $\mathcal{S}_i, \ \forall i$. By minimizing this energy, we force the images of the $P_i$ to have few non-zero entries.
\textcolor{red}{Together with the choice of the sum as aggregation operator for the $\mathcal{S}_i, \ i=1 \ldots k $, this imposes the structure of a Cartesian product space on $\mathcal{Z}=\mathcal{S}_1 \times \ldots \mathcal{S}_k$, as we prove in the Appendix.}
%(the minimum required to encode information).
% Moreover, it allows us to aggregate the vectors in each latent space via the vector sum, as this operation coincides with the concatenation of the latent vectors, omitting the zero entries.
Furthermore, this loss is equivalent to setting a constraint on the size of the learned latent space, which is  needed to prevent unbounded optimization.
%forse potremmo provare qualcuno di questi claim

\textbf{Training strategy.}
We train in two phases. First, we aim to learn the global structure of the manifold, relying solely on the reconstruction loss. Once we have given a structure to the latent space, we can start factorizing it into subspaces as the other terms in the loss enter the minimization.
It is crucial that the space $\mathcal{Z}$ will continue to change as the other losses take over, since decoupling completely the reconstruction from the disentanglement could lead to not-separable spaces.
The regularization parameters $\beta_i, i=1,2,3$  are used to impose this behavior. For further details, we refer to the \sout{Supplementary}\textcolor{red}{Appendix}.

\textbf{Regularization loss.}
In the first phase of the training, where we mainly aim to achieve good reconstruction quality, \textcolor{red}{we have no guarantees that the information of the factors of variation is correctly spread among the subspaces, and this may lead to a single subspace encoding multiple factors of variation. To avoid this problem,  we introduce a penalty that ensures the choice of the oracle $\tilde{\mathcal{O}}$ is equally distributed among the subspaces. }\sout{we must ensure that the information is equally spread among the subspaces, i.e. the distances within each subspace must not \textcolor{red}{be} altered too much.}
In practice, the indicator variables in Eq.\ref{Eq:dist_loss} are approximated in each batch of $N$ samples with a matrix $\mathbf{A}$ of dimensions $N\times k$, obtained by applying a weighted softmax to the distance matrix of pairs in each of the $k$ subspaces. This matrix acts as a differentiable mask that implements the oracle $\tilde{\mathcal{O}}$ in practice. In the reconstruction phase, we activate a penalty on the matrix:
\begin{align}
    \mathcal{L}_{reg}= \sum_{j=1}^k \left(\frac{1}{N}\sum_{n=1}^N \mathbf{A}_{n,j}- \frac{1}{k} \right)^2\,,
\end{align}
which ensures that the selection of each subspace has the same probability.

\section{Experimental results}

%experimental settings
%datasets
%metriche
%results
%comparison
%analysis
    %robustness to k
    %benefits of higher dimensioal subspaces
        %faust
        %dsprites

To validate our model we perform experiments on both synthetic and real datasets. In the synthetic experiments, observations are generated as a deterministic function of the known factors of variation. For the real settings, since the data are collected directly from real observations, the parametric function generating the data is not known, and could have a different number of parameters than the number of factors that we aim to disentangle.
To put ourselves in a setting comparable to the competing methods, in all our experiments we use a latent space of dimension $d=10$, unless otherwise specified, and $k=10$ latent subspaces.

\textbf{Synthetic data.}
We adopted 4 widely used synthetic datasets in order to evaluate the effectivness of our method, namely \emph{DSprites} \cite{Higgins2017}, \emph{Shapes3D} \cite{kim2019}, \emph{Cars3D} \cite{Reed:2015}, \emph{SmallNORB} \cite{LeCUn:2004} . All these datasets contain images that are parametrically rendered based on some known factors of variation, the goal is thus to obtain a disentangled latent space that reflects these parameters. We resized the input images to a dimension of $64 \times 64$ pixels. During training, we give as input to our method two images which differ only by a single randomly sampled factor of variation, and this is the {\em only} assumption that we make on the input data. We do not provide any information on which of the factors is changing between the two images.
\textcolor{red}{We show quantitative results on these datasets in Tables \ref{tab:comparison-shapes3d},\ref{tab:comparison-dsprites},\ref{tab:comparison-cars3d},\ref{tab:comparison-smallnorb} comparing our method to the state-of-the-art model  Ada-GVAE of \cite{locatello2020}, in terms of the metrics described below.}

\textbf{Real data.}
As a benchmark for real data, we test our method on FAUST \cite{Bogo:2014}, a dataset composed by 100 3D human scans of 10 subjects in 10 different poses. Differently from the synthetic case, the data does not derive from a parametric model. This dataset is particularly challenging, since each of the two factors of variation (pose and identity) is difficult to embed in a 1-dimensional space; indeed, human parametric models define pose and identity using dozens of parameters \textcolor{red}{\cite{SMPL:2015}}.  To speedup the training, we remeshed each shape to 2.5K points.

%To generate the sampled pairs of observations, we proceed as follows: we randomly sample one vector of the true generative factors. Then we sample an index $i$,  which will correspond to the factor of variation, and  we sample a new value for it and substitute it in the vector, keeping the other factors fixed.

\textbf{Evaluation metrics.}
To quantitatively evaluate our experiments we adopt a set of evaluation metrics widely used in the disentanglement literature: namely, the $\beta$-VAE score and the Factor VAE score, which measure disentanglement as the accuracy of a linear classifier that predicts the index of a fixed factor of variation and  a majority vote classifier respectively \cite{Higgins2017,kim2019}; the DCI disentanglement score,  which measures the entropy of the distribution on the single dimension of the representation vector \cite{Eastwood:2018}; and the mutual information gap (MIG) \cite{chen:2019} which for each dimension of the representation vector, first measures the mutual information w.r.t. the other dimensions and then takes the gap between the highest and second coordinate. 
Note that the aforementioned metrics have been defined to measure the disentanglement in a variational framework, wherein the factors of variation are assumed to be one-dimensional.
%
% \begin{minipage}[t]{\linewidth}
\begin{table}[t!]
\caption{Median disentanglement scores on Shapes3D.}
\label{tab:comparison-shapes3d}
\begin{center}
\begin{small}
% \begin{sc}
\begin{tabular}{l|cccc}
\toprule
& $\beta$-VAE & $\beta$-TCVAE & Ada-GVAE  & Ours   \\
\sc{Metric}\\
\midrule
BetaVAE    &  98.6\% &  99.8\%   &  \bf{100}\%   &  \bf{100}\%  \\
FactorVAE  &  83.9\% &  86.8\%   &  \bf{100}\%   &  \bf{100}\%  \\
DCI Disent.&  58.8\% &  70.9\%   &  94.6\%  &  \bf{99.0}\%  \\
MIG        &  22.0\% &  27.1\%   &  56.2\%  &  \bf{63.7}\%  \\
MIG-PCA    &    -    &   -       &     -    &  73.5\%  \\
 MIG-KM  &    -    &   -       &     -    &  69.2\%  \\
\end{tabular}
% \end{sc}
\end{small}
\end{center}
%\vskip 0.1in
% \end{table}
%

% \begin{table}[h!]
\caption{Median disentanglement scores on DSprites.}
\label{tab:comparison-dsprites}
\begin{center}
\begin{small}
% \begin{sc}
\begin{tabular}{l|cccc}
\toprule
& $\beta$-VAE & $\beta$-TCVAE & Ada-GVAE  & Ours   \\
\sc{Metric}\\
\midrule
BetaVAE    &  82.3\% &  86.4\%   &  92.3\%   &  \bf{98.9}\%  \\
FactorVAE  &  66.0\% &  73.6\%   &  84.7\%   &  \bf{88.5}\%  \\
DCI Disent.&  18.6\% &  30.4\%   &  47.9\%  &  \bf{49.7}\%  \\
MIG        &  10.2\% &  18.0\%   &  \bf{26.6}\%  &  25.7\%  \\
MIG-PCA    &    -    &   -       &     -    &  20.9\%  \\
 MIG-KM  &    -    &   -       &     -    &  17.9\%  \\
\end{tabular}
% \end{sc}
\end{small}
\end{center}
%\vskip 0.1in
% \end{table}
%
% \begin{table}[h!]
\caption{Median disentanglement scores on Cars3D.}
\label{tab:comparison-cars3d}
\begin{center}
\begin{small}
% \begin{sc}
\begin{tabular}{l|cccc}
\toprule
& $\beta$-VAE & $\beta$-TCVAE & Ada-GVAE  & Ours   \\
\sc{Metric}\\
\midrule
BetaVAE    &  100\% &  100\%   &  \bf{100}\%   &  \bf{100}\%  \\
FactorVAE  &  87.9\% &  90.2\%   &  \bf{90.2}\%   &  89.9\%  \\
DCI Disent.&  22.5\% &  27.8\%   &  \bf{54.0}\%  &  48.9\%  \\
MIG        &  8.8\% &  12.0\%   &  15.0\%  &  \bf{25.9}\%  \\
MIG-PCA    &    -    &   -       &     -    &  20.6\%  \\
MIG-KM  &    -    &   -       &     -    &  17.8\%  \\
\end{tabular}
% \end{sc}
\end{small}
\end{center}
% \end{table}
%
% \begin{table}[h!]
\caption{Median disentanglement scores on SmallNorb.}
\label{tab:comparison-smallnorb}
\begin{center}
\begin{small}
% \begin{sc}
\begin{tabular}{l|cccc}
\toprule
& $\beta$-VAE & $\beta$-TCVAE & Ada-GVAE  & Ours   \\
\sc{Metric}\\
\midrule
BetaVAE    &  74.0\% &  76.5\%   &  87.9\%   &  \bf{99.2}\%  \\
FactorVAE  &  49.5\% &  54.2\%   &  55.5\%   &  \bf{88.2}\%  \\
DCI Disent.&  28.0\% &  30.2\%   &  33.8\%  &  \bf{49.4}\%  \\
MIG        &  21.4\% &  21.0\%   &  25.6\%  &  \bf{25.8}\%  \\
MIG-PCA    &    -    &   -       &     -    &  20.8\%  \\
MIG-KM  &    -    &   -       &     -    &   17.8\% \\
\end{tabular}
% \end{sc}
\end{small}
\end{center}
%%\vskip 0.1in
\end{table}
% \end{minipage}
%
This is not the case in our method, where each disentangled subspace is {\em not} required to be one-dimensional. Nevertheless, we can still apply these metrics in the aggregated latent space. This allows us to perform a direct comparison with methods working in the variational setting.

%Since our framework generalizes to multi dimensional latent subspaces, to make a fair comparison we put ourselves in the one dimensional context, aggregating the dimensionalities within the subspace.
%In particular, the mutual information gap measures the normalized gap of mutual information between the first and the second highest coordinates in the latent representantion.
We also introduce two new metrics, which adapt the Mutual Information Gap (MIG) score to a multi-dimensional space. 
The former, which we denote by MIG-PCA, projects the latent subspaces onto the principal axis of variation through PCA \cite{Pearson;1901}. This way we can directly reconduct ourselves to the variational setting, where each factor is encoded by just one dimension of the latent space, and use directly the MIG score.
The latter, denoted by MIG-KM, is based on the use of k-means in the multi-dimensional subspace to derive a one-dimensional discrete random variable. In practice, on each subspace, we extract $b$ centroids using k-means on the considered evaluation samples. Each sample is then assigned to one of these $b$ centroids, thus deriving a probability distribution over the centroids. The Mutual Information is then computed  between this distribution and the distribution of the ground-truth labels in order to derive the MIG score. This latter metric is particularly important in the real data scenario, where each factor of variation is likely to be multi-dimensional, while MIG-PCA still assumes the factor of variation to be represented in a linear subspace.
%
%These adaptations are necessary to fairly compare our method with the other work, but we stress that information is inherently lost in the dimensionality reduction process of $MIG-PCA$. As k-means is adaptable to arbitrary dimensions the method fits our needs. The proposed metrics mitigates the loss information as testified by our experimental results reported in the tables, where they outperform the MIG metric computed directly on the product of the subspaces.

\textbf{Implementation details.}
We performed our experimental evaluation on a machine equipped with an NVIDA RTX 2080ti, within the Pytorch framework. The architecture for encoder and decoder on images are convolutional, with the same exact settings as \cite{locatello2020}. For the experiments on FAUST we used a PointNet \cite{qi2017} architecture and a simple MLP as a decoder. Detailed information on the architecture, including the hyperparameter choice, are reported in the \sout{Supplementary}\textcolor{red}{Appendix}.

\begin{figure}[t]
    \centering
    \includegraphics[width=\linewidth]{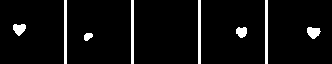}
    \vskip 0.03in
    \includegraphics[width=\linewidth]{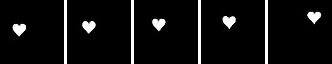}
    \caption{Interpolation example on the prevalent latent dimension / subspace of the changing factor on a pair of images from {\em Dsprites}; we compare Ada-GVAE (top) with our method (bottom). To highlight the importance of having multi-dimensional disentangled subspaces, we trained the network by aggregating the translation factors along the {\em x} and {\em y} dimensions into a single one, making it multi-dimensional by construction. Even if reaching a similar MIG score (Ada-GVAE: 50.2\%, Ours: 49.3\%), the interpolation obtained with our method is better behaved.}
    \label{fig:dsprites}
\end{figure}

\begin{table}[t]
%\vskip -0.4cm
\caption{Comparison of disentanglement metrics and reconstruction score at varying latent space dimensions on the FAUST dataset. We trained our model by fixing the number of latent subspaces to $2$, while allowing different dimensions of the global latent space to $2,4,8,16$.}
\label{sample-table}

\begin{center}
\begin{small}
% \begin{sc}
% \begin{tabular}{p{0.16\linewidth} | p{0.1\linewidth} p{0.1\linewidth} p{0.1\linewidth} p{0.1\linewidth} p{0.1\linewidth}}
\begin{tabular}{lcccc}
\toprule
&\multicolumn{4}{c}{MIG}\\
\midrule
 & 2 & 4 & 8 & 16 \\
\midrule
\multicolumn{1}{l|}{Ada-GVAE} & 15.1\%& \bf{11.1}\%& \bf{4.6}\%& 2.3\%\\
\multicolumn{1}{l|}{Ours}     & \bf{18.2}\% & 4.39\% & 3.08\%& \bf{2.68}\%\\
\midrule
&\multicolumn{4}{c}{MIG-KM}\\
\midrule
& 2 & 4 & 8 & 16 \\
\midrule
\multicolumn{1}{l|}{Ours} & - & 10.6\% & 10.3\%& 9.5\%\\
\midrule
 &\multicolumn{4}{c}{Reconstruction error}\\
 \midrule
& 2 & 4 & 8 & 16 \\
\midrule
\multicolumn{1}{l|}{Ada-GVAE} & 7.22e-3& 5.11e-3& 3.39e-3& 3.30e-3\\
\multicolumn{1}{l|}{Ours} & \bf{2.56e-3} & \bf{2.04e-3} & \bf{1.24e-3} & \bf{1.26e-3}\\
\bottomrule
\end{tabular}
% \end{sc}
\end{small}
\end{center}

\end{table}

\begin{table}[h!]
\caption{Performance comparison of our method with and without fixing the number of changing factors to exactly one. We report median results on the {\em Shapes3D} dataset.}
\label{tab:variablek}

\begin{center}
\begin{small}
% \begin{sc}
% \begin{tabular}{p{0.16\linewidth} | p{0.1\linewidth} p{0.1\linewidth} p{0.1\linewidth} p{0.1\linewidth} p{0.1\linewidth}}
\begin{tabular}{p{0.16\linewidth} | ccccc}
\toprule
& \multirow{2}{0.1\linewidth}{Beta VAE}& \multirow{2}{0.1\linewidth}{DCI Dis.}& \multirow{2}{0.1\linewidth}{MIG}& \multirow{2}{0.1\linewidth}{MIG-PCA}& \multirow{2}{0.1\linewidth}{MIG-KM}\\
%  & \multirow{2}{BetaVAE}  & \multirow{2}{DCI Dis.} & \multirow{2}{MIG} & \multirow{2}{MIG-PCA} & \multirow{2}{MIG-KM} \\
\#factors \\
\midrule
One      & 100\% & 99.0\% &  63.7\% & 73.5\% &  69.2\%  \\
Variable & 98.9\% & 94.9\% & 62.3\% &  70.5\%& 66.9\% \\
\bottomrule
\end{tabular}
% \end{sc}
\end{small}
\end{center}
\end{table}

\subsection{Analysis}
\textcolor{red}{In the following we analyze the core properties of our approach.}

\begin{table*}[t]
\caption{Average standard deviation on the latent subspaces corresponding to Fig.~\ref{fig:latent_space}. The lower scores correspond to collapsed subspaces.}
\label{tab:avgstd}

\begin{center}
\begin{small}
\begin{tabular}{ l|cccccccccc}
\toprule
{\textbf{Subspace}} & $ \mathcal{S}_1$ &$\mathcal{S}_2$ & $\mathcal{S}_3$ & $\mathcal{S}_4$ & $\mathcal{S}_5$ & $\mathcal{S}_6$ & $\mathcal{S}_7$ & $\mathcal{S}_8$ & $\mathcal{S}_9$ & $\mathcal{S}_{10}$  \\
 %\midrule
 {\textbf{Average std}} & 0.061 & 0.043  & 0.063  & 0.026  & 2.6e-5  & 2.1e-5  & 1.5e-5  & 0.058  & 0.063  & 2.1e-5 \\
\bottomrule
\end{tabular}

\end{small}
\end{center}
\end{table*}

\begin{figure*}[t]
    \centering
    \includegraphics[width=0.9\linewidth]{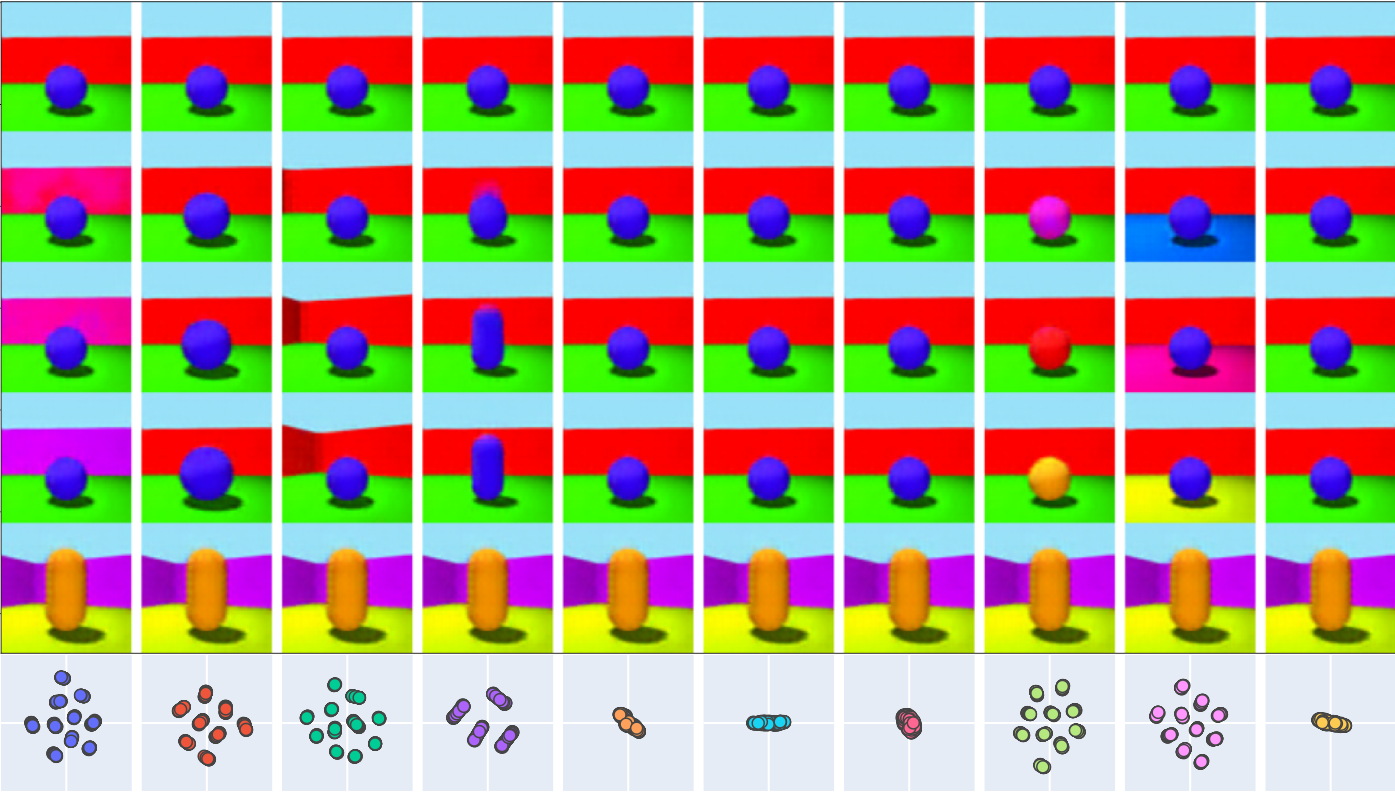}
    \caption{Analysis of the latent subspaces at training convergence on the \emph{Shapes3D} dataset. We project each latent subspace (ordered by column number) on a 2-dimensional space, using t-SNE (last row). On the rows we show the interpolation (rows 2,3) between two samples which differ on all factors of variation (rows 1,4). We show that the latent subspaces (columns 1,2,3,4,8,9) which encode information about the generative factors (wall hue, size, camera angle, shape, object hue, floor hue) are correctly characterized by the latent subspace projections, and the number of clusters is almost always equal to the number of possible values that the generative factor can assume. For example, the number of clusters in the first subspace corresponds to the ten possible color hue values for the wall, and the number of clusters in the 4th subspace corresponds to the number of possible shapes in the dataset. The latent spaces which do not encode any information (columns 5,6,7,10) collapse to one point (represented by a single, small cluster in the projection), as confirmed by the quantitative evaluation in Table~\ref{tab:avgstd}. }
    \label{fig:latent_space}
\end{figure*}
\textbf{Inferring the number of latent subspaces.}
We stress that our method is tailored for learning the number of subspaces, and therefore the number of generative factors, as well as the dimensionality of each of them. The only hyperparameter that needs to be set is the dimensionality of the latent space $\mathcal{Z}$ to be factorized. To infer the number of subspaces we over-estimate this parameter, with the desirable consequence that the network will activate only some of them, corresponding to the number of ground truth generative factors. The remaining ones will collapse to a point, as we can observe in the qualitative example\sout{ of Figure~\ref{fig:3dshapes}.} \textcolor{red}{of Figure~\ref{fig:latent_space} accompanied by the quantitative result in Table~\ref{tab:avgstd}.}
% To show that our method is capable of inferring the correct number of subspaces, and also to show how properly they are characterized, we perform a qualitative and quantitative analysis on the latent subspaces (Figure \ref{fig:latent_space}). We project each latent subspace (approximated by randomly sampling 512 elements in the dataset) onto a 2-dimensional subspace, using the dimensionality reduction technique of (t-SNE)~\cite{vandermaaten08}. We show which subspace encodes information about which factor by counting the number of clusters that are formed by t-SNE. At convergence, the number of clusters will correspond loosely to the number of possible values that the generative factor can assume (e.g. if the true distribution of the generative factor  is discrete and  10-dimensional, we expect to count 10 clusters). We can also see that our algorithm is capable of inferring the correct number of subspaces from an overestimate, this reflects in the fact that when a factor does not encode any information it collapses to a single cluster. We accompany each latent subspace projection plot with an interpolation of pair of samples, which differ on each factor of variation on each of the subspace. These results are supported by the quantitative evidence in Table~\ref{tab:avgstd}, where we measure the average standard deviation in each subspace on the same random sampling plotted in 2 dimensions; we show that there is a strong gap between variance in the subspaces that encode information, and those that are collapsed.

\textbf{Expressiveness of higher dimensional latent subspaces.}
We further report experimental evidence that having multi-dimensional subspaces allows to obtain a better disentangled representation.
We show this on the challenging FAUST dataset \cite{Bogo:2014}, where the factors of variation are not reducible to just one dimension.
As a test for our method, we confront ourselves with the challenging task of separating pose and style for human 3D models in different poses and identities.

\begin{figure*}[t]
\centering
\begin{overpic}[trim=0 -8cm 0 0,clip,width=0.1\linewidth]{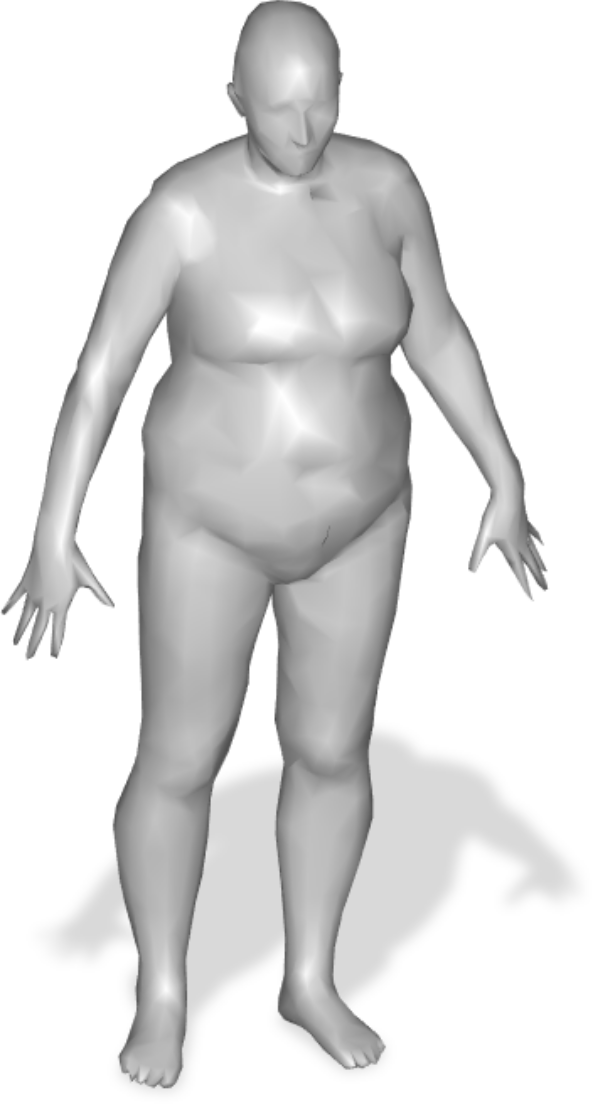}
\put(14,14){$x_1$}
\end{overpic}
\hfill
\begin{overpic}[ trim=-3cm -3cm 0 -3cm, clip, width=0.35\linewidth]{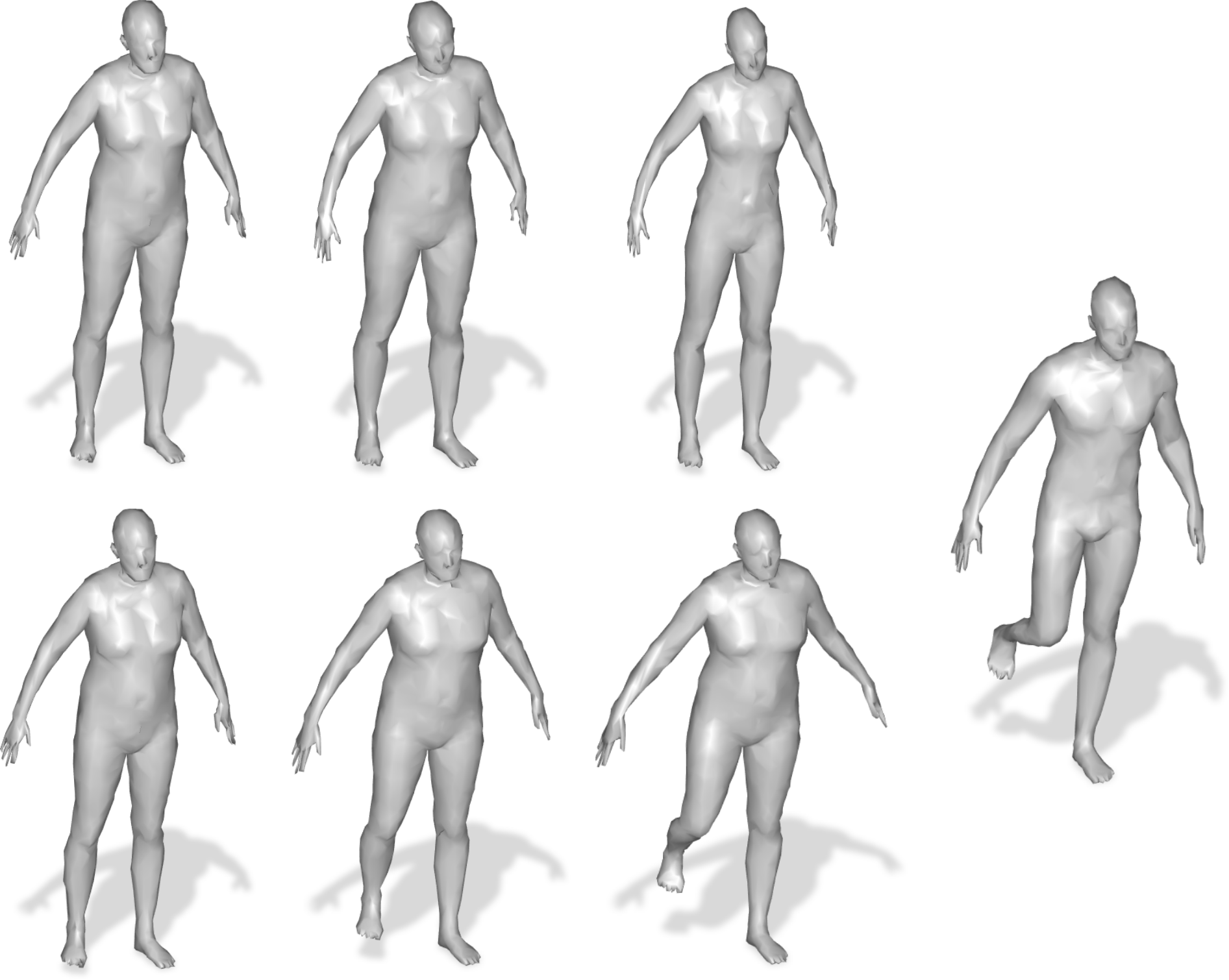}
\put(-2,20){\small\rotatebox[]{90}{\textit{int}($s_1^1$,$s_2^1$,$\alpha$)}}
\put(-2,60){\small\rotatebox[]{90}{\textit{int}($s_1^2$,$s_2^2$,$\alpha$)}}
\put(9,82){\small$\alpha=0$}
\put(30,82){\small$\alpha=0.5$}
\put(56,82){\small$\alpha=1$}
\put(88,63){$\tilde{x}_2$}
\put(7,5){\line(1,0){90}}
\put(33,0){$k=2, d=2$}
\end{overpic}
\hfill
\begin{overpic}[ trim=-3cm -3cm 0 -3cm, clip, width=0.35\linewidth]{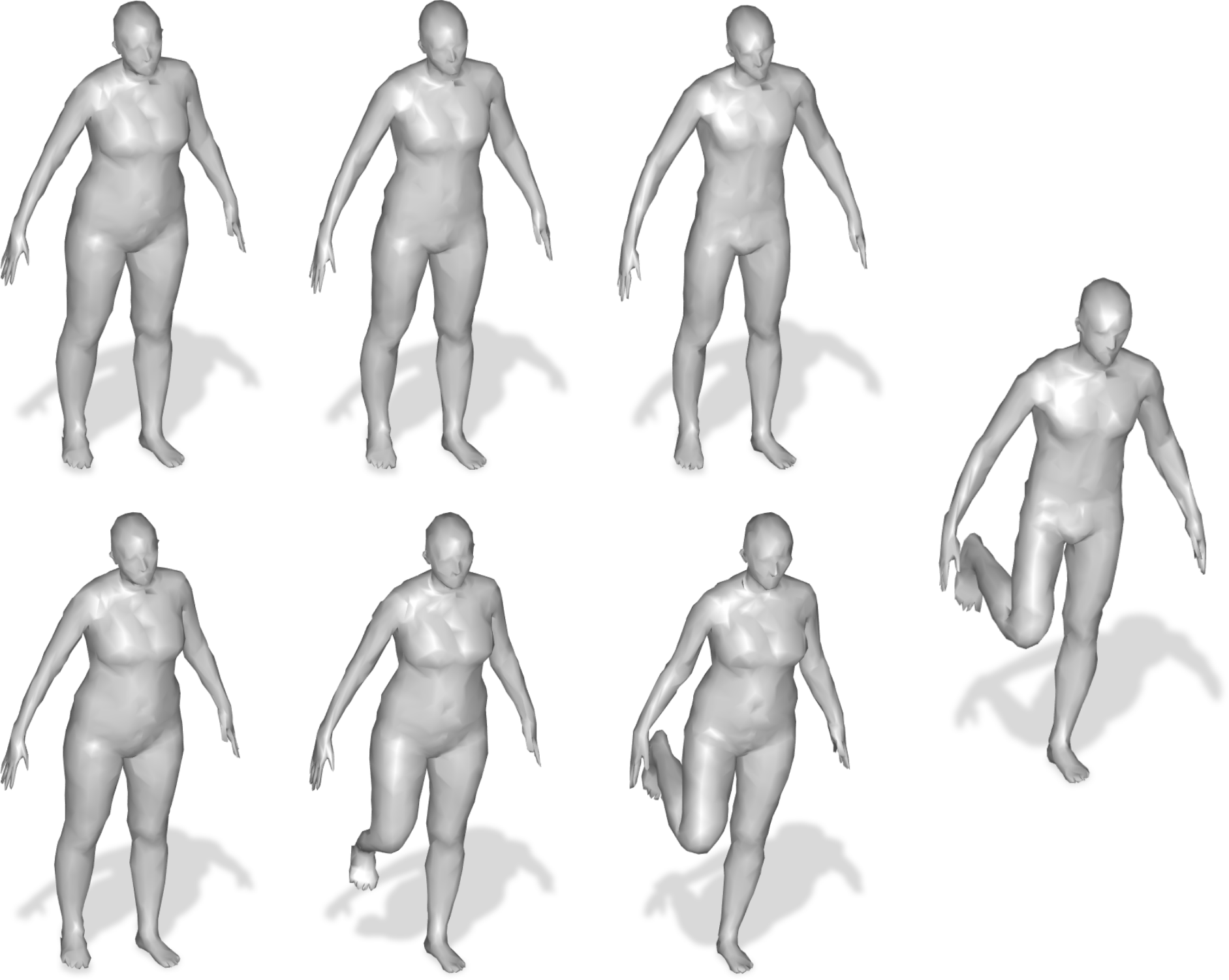}
\put(-2,20){\small\rotatebox[]{90}{\textit{int}($s_1^1$,$s_2^1$,$\alpha$)}}
\put(-2,60){\small\rotatebox[]{90}{\textit{int}($s_1^2$,$s_2^2$,$\alpha$)}}
\put(9,82){\small$\alpha=0$}
\put(30,82){\small$\alpha=0.5$}
\put(56,82){\small$\alpha=1$}
\put(87,63){$\tilde{x}_2$}
\put(7,5){\line(1,0){90}}
\put(33,0){$k=2, d=16$}
\end{overpic}
\hfill
\begin{overpic}[trim=0 -8cm 0 0,clip,width=0.092\linewidth]{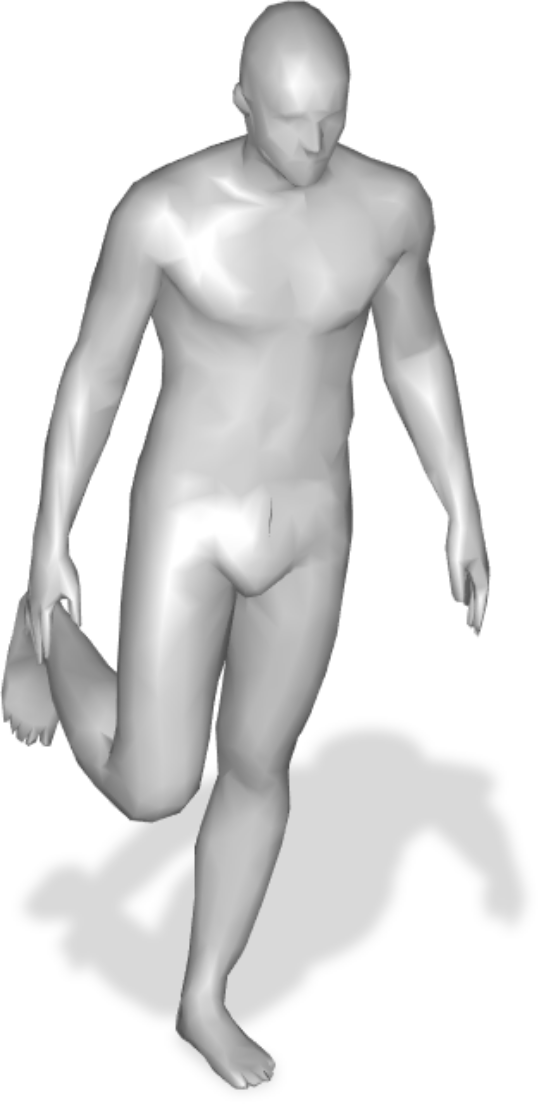}
\put(13,14){$x_2$}
\end{overpic}
\caption{Disentangled interpolation on a pair of FAUST shapes with different dimensions of the latent space. For each of the two latent space dimensions ($d=2$ on the left and $d=16$ on the right) we encode the two input shapes $x_1$ and $x_2$ as subspace latent vectors $(s_1^1, s_1^2)$ for $x_1$ and $(s_2^1, s_2^2)$ for $x_2$. We then interpolate separately the 2 latent subspaces between the two shapes, and obtain the decoded shapes $g(aggr( s_1^1, int(s_1^2,s_2^2,\alpha)))$ (top row) and $g(aggr(int(s_1^1,s_2^1,\alpha),s_1^2 ))$ (bottom row), with $int(a,b,\alpha)=(1-\alpha) a + \alpha b$. We can see how, even if both configurations are able to achieve the disentanglement between pose and style of the subjects, the smaller latent space ($d=2$) is not able to encode the finer shape details, resulting in smoother and less accurate reconstructions.
\label{fig:faust_interp}}
\end{figure*}

In Table \ref{sample-table} we ran this experiment while growing progressively the dimensionality of the latent space $\mathcal{Z}$ from 2 to 16.
Crucially, we show that adopting progressively larger multi-dimensional subspaces does not degrade the disentanglement quality as measured by the MIG-KM score, while the reconstruction error goes down faster than with one dimension. The standard MIG score shows that in the one-dimensional case the performance of disentanglement degrades. A qualitative example is reported in Figure~\ref{fig:faust_interp}. A qualitative comparison  with \cite{locatello2020} on a different dataset is also shown in Figure~\ref{fig:dsprites}.
 
 We remark here that the constraint of observing data sampled in pairs is a reasonable assumption, because data come often in this form in the acquisition stage; examples include temporal consistency of video frames~\cite{Lai:2018}, object identity consistency of a rendered object from different camera angles, etc.

\begin{figure*}[!h]
\begin{minipage}{0.7\linewidth}
\begin{center}
\begin{small}
% \begin{sc}
% \begin{tabular}{p{0.16\linewidth} | p{0.1\linewidth} p{0.1\linewidth} p{0.1\linewidth} p{0.1\linewidth} p{0.1\linewidth}}
\begin{tabular}{l|cccccc}
\toprule
& \multirow{2}{*}{Beta VAE}& \multirow{2}{0.08\linewidth}{Factor VAE}& \multirow{2}{*}{DCI Dis.}& \multirow{2}{*}{MIG}& \multirow{2}{*}{MIG-PCA}& \multirow{2}{*}{MIG-KM}\\
%  & \multirow{2}{BetaVAE}  & \multirow{2}{DCI Dis.} & \multirow{2}{MIG} & \multirow{2}{MIG-PCA} & \multirow{2}{MIG-KM} \\
Loss terms \\
\midrule
Complete      & \textbf{100\%}& \textbf{100\%f} & 96.1\% &  \textbf{81.0\%} & \textbf{86.0\%} &  \textbf{80.3\%}  \\
W/o $\mathcal{L}_{reg}$ & 99.5\% & 99.8\% & 77.2\% & 48.2\% &  58.6\%& 51.5\% \\
W/o $\mathcal{L}_{cons}$    & \textbf{100\%} & 97.1\% & \textbf{97.0\%} &  65.4\% & 81.0\% &  71.0\%  \\
W/o $\mathcal{L}_{reg},\mathcal{L}_{cons}$ & 93.4\% & 87.9\% & 78.2\% & 56.2\% &  55.5\%& 53.8\% \\
\bottomrule
\end{tabular}
% \end{sc}
\end{small}
\end{center}
\vfill
\end{minipage}
\hfill
\begin{minipage}{0.23\linewidth}
% \begin{figure}
\begin{overpic}[trim=0cm -1.4cm 0cm -1.3cm, clip,width=1\linewidth]{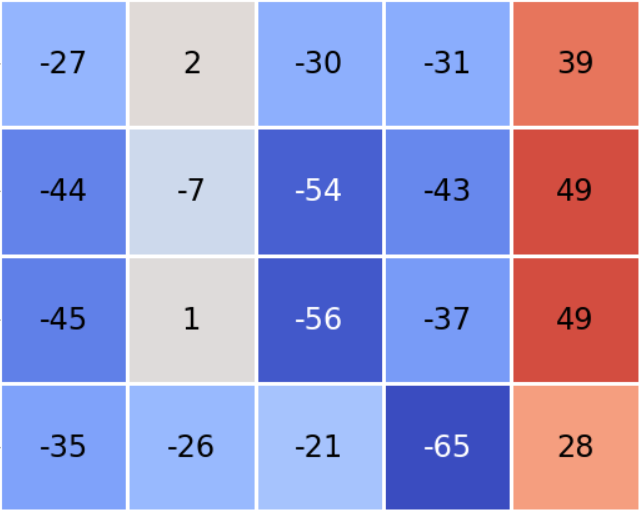}
    \put(20, 92){Rank correlation }
    \put(83, -0){\small\rotatebox[]{-45}{$\mathcal{L}_{rec}$}}
    \put(64, -0){\small\rotatebox[]{-45}{$\mathcal{L}_{reg}$}}
    \put(44, -1){\small\rotatebox[]{-45}{$\mathcal{L}_{cons}$}}
    \put(24, 0.5){\small\rotatebox[]{-45}{$\mathcal{L}_{dis}$}}
    \put(3, -){\small\rotatebox[]{-45}{$\mathcal{L}_{spar}$}}

    \put(-14, 77){\tiny MIG}
    \put(-14, 60){\tiny MIG-}\put(-14, 55){\tiny PCA}
    \put(-14, 41){\tiny MIG-}\put(-13, 36){\tiny KM}
    \put(-14, 19){\tiny DCI}
\end{overpic}
    %\caption{Rank correlation between the different loss terms and disentanglement scores.\label{fig:correlation}}
% \end{figure}
\end{minipage}
\captionlistentry[table]{A table beside a figure}
\label{tab:ablation}
\captionsetup{labelformat=andtable}
\caption{Analysis of the importance of the adopted loss terms on the {\em Shapes3D} dataset. The table (\textbf{left}) reports ablation results; these show the disentanglement scores obtained by training our model with a fixed initialization seed, and removing some of the loss terms. We can see how all the losses contribute to achieve a better disentanglement. A similar conclusion can be drawn looking at the Rank correlation matrix (\textbf{right}), showing the rank correlation between the losses and the disentanglement scores obtained after training 10 models. All the disentanglement losses (first 4 columns) are negatively correlated with disentanglement scores. The disentanglement loss $\mathcal{L}_{dis}$ seems to be less correlated with respect to the others, which is due to the fact that it always reaches a similar low value in all the runs. 
\label{fig:ablation}}
\end{figure*}

\textbf{Robustness to simultaneous changes of multiple factors.}
Although the proposed framework assumes that the observed pairs change in only one factor of variation, we show that, in practice, our method performs well also in settings where more than one factor may randomly vary simultaneously. 
The only assumption we make is that at least one factor is shared among the elements of  the sampled pairs. To build the pairs of observations we follow the same process as before, where we sample a random number between $1$ and $k-1$ transformed factors between the elements of the sampled pair.
Our method proves to be robust to this challenging setting, with performance comparable to the case with one fixed factor of variation, as we show in Table~\ref{tab:variablek}.

\textbf{Ablation study.}\label{sec:ablation}
We justify the importance of the consistency loss and the distribution loss by running some ablation experiments. We fix a random seed and run experiments on the \emph{Shapes3D} dataset with a complete run, a run without the regularization loss, a run without the consistency loss, and a run without both. The importance of using all our loss terms is highlighted in Table \ref{tab:ablation}, where we report the $\beta$-VAE, Factor-VAE, DCI and MIG scores for these setups. This is further supported by the rank correlation matrix between losses and disentanglement scores shown in Figure~\ref{fig:ablation}.

\section{Conclusions}
% In this paper, we introduce a new definition of disentanglement based on the notion of metric space, which generalizes existing ones. Moreover, we pose the following two questions: (i) Can disentangled representations be learned in this setting by observing solely non i.i.d. samples?  (ii) Can these representation benefit for allowing higher dimensionality of the subspaces $> 1$?
% We answer positively, providing a simple recipe to compute disentangled representations in practice. The proposed approach achieves state-of-the-art performance on different datasets, and can be easily adapted to many standard neural network models.
In this paper, we introduced a new definition of disentanglement based on the notion of metric space, which generalizes existing approaches. Relying on this definition, we proposed a simple recipe to compute disentangled representations in practice, in a weakly supervised setting. 
In contrast with previous work, we  demonstrated that disentanglement representations benefit from the choice of modelling generative factors as possibly multi-dimensional subspaces, especially when the true factors of variation live in a space with dimension greater than $1$.
%We prove that our paradigm could also apply when the dimensionality of the subspaces, where the factor of variations live is greater than $1$.
In many cases, the proposed solution outperforms state-of-the-art competitors on synthetic datasets and more challenging real-world scenarios. Moreover, our method can be easily adapted to many standard neural network models as a plugin module.

% \textbf{Limitations.}
\textbf{Future work.}
Exploiting the possible contribution of the proposed model in existing autoencoders is a natural direction raised by this paper.
In the future, we aim to investigate possible alternative constraints to incorporate additional properties in the latent space or its factorization. %W%e could, for example, explicitly require that the latent subspaces 
\section*{Acknowledgements}
\textcolor{red}{The authors are supported by the ERC Starting Grant No. 802554 (SPECGEO) and the MIUR under grant ``Dipartimenti di eccellenza 2018-2022'' of the Department of Computer Science of Sapienza University.}

\bibliography{bibliography}
\bibliographystyle{icml2021}

%FOR THE ARXIV

\clearpage
\onecolumn
\appendix
\begin{fleqn}

\section{Implementation details}
For the experiments on the datasets \emph{DSprites} \cite{Higgins2017}, \emph{Shapes3D} \cite{kim2019}, \emph{Cars3D} \cite{Reed:2015}, \emph{SmallNORB} \cite{LeCUn:2004}, we implement a simple convolutional architecture for both the encoder and the decoder. We report the detailed parameters in Table \ref{tab:conv}, where $d$ refers to the dimensionality of the latent space $\mathcal{Z}$, which bounds the maximum dimensionality of each of the $k$ latent subspaces $\mathcal{S}_1 \ldots \mathcal{S}_k$. The architecture of the nonlinear projectors $P_i$ is described in Table \ref{tab:proj}. 
For the {\em FAUST} dataset we employ a PointNet \cite{qi2017} based architecture for the encoder and a simple MLP for the decoder. Details are reported in Table \ref{tab:pointnet}

\begin{table}[h!]
\caption{Convolutional architecture used in image datasets.}
\label{tab:conv}

\begin{center}
\begin{small}
\begin{tabular}{ cc}
\toprule
 {\textbf{Encoder}}& {\textbf{Decoder}}\\
\midrule
 Input : $64 \times 64 \times$ number of channels & Input : $\mathbb{R}^d$ \\
$4 \times 4$conv, $32$ ReLU, stride $2$, padding $1$ & FC, $256$, ReLU\\
$4 \times 4$conv, $32$ ReLU, stride $2$, padding $1$ & FC, $256$, ReLU\\
$4 \times 4$conv, $64$ ReLU, stride $2$, padding $1$ & FC, $64\times 4 \times 4$, ReLU\\
$4 \times 4$conv, $64$ ReLU, stride $2$, padding $1$ & $4 \times 4$upconv, $64$ ReLU, stride $2$, padding 1\\
FC, $256$, ReLU & $4 \times 4$upconv, $32$ ReLU, stride $2$, padding 1\\
FC, $d$ & $4 \times 4$upconv, $32$ ReLU, stride $2$, padding 1\\
- & $4 \times 4$upconv, number of channels, stride $2$, padding 1\\
\bottomrule
\end{tabular}

\end{small}
\end{center}
\end{table}

\begin{table}[h!]
\begin{minipage}{.73\linewidth}
\caption{PointNet - MLP architecture used in FAUST dataset.}
\label{tab:pointnet}
\begin{center}
\begin{small}
\begin{tabular}{ cc}
\toprule
 {\textbf{Encoder}}& {\textbf{Decoder}}\\
\midrule
 Input : $ 2500 \times$ 3 & Input : $\mathbb{R}^d$ \\
$1 \times 1$conv, $32$, BatchNorm, ReLU,& FC, 1024, LeakyReLU\\
$1 \times 1$conv, $128$, BatchNorm, ReLU,& FC, 2048, LeakyReLU\\
$1 \times 1$conv, $256$, BatchNorm, ReLU,& FC, $2500\times 3$, ReLU\\
$1 \times 1$conv, $512$,&-\\
MaxPooling,&-\\
FC, $512$, BatchNorm, ReLU,& -\\
FC, $256$, BatchNorm, ReLU,& -\\
FC, $128$, BatchNorm, ReLU,& -\\
FC, $ d$,& -\\
\bottomrule
\end{tabular}
\end{small}
\end{center}
\end{minipage}%
\begin{minipage}{.15\linewidth}
\caption{Projectors architecture.}
\label{tab:proj}
\begin{center}
\begin{small}
\begin{tabular}{c}
\toprule
\textbf{P} \\
\midrule
Input : $ \mathbb{R}^d$   \\
FC $ d$, ReLU \\
FC $ d$ \\
\bottomrule
\end{tabular}
\end{small}
\end{center}
\end{minipage}
\end{table}

\subsection{Experimental settings}
For the comparisons with \cite{locatello2020} and its top performer model Ada-GVAE presented in Tables 1-4 in the main paper, we set the dimensionality $d$ of the latent space $\mathcal{Z}$ to 10, and the number of subspaces $k$ to 10. This puts us in a setting that is as close as possible to \cite{locatello2020}, where the latent space is 10-dimensional and the subspaces are 1-dimensional by construction. For all the quantitative experiments we trained 5 times the same model with different random seeds, and report the median results on each dataset. A summary of the hyperparameters are in Table~\ref{tab:hyperparams}.

\begin{minipage}{1\linewidth}
\begin{minipage}{.48\linewidth}
\begin{center}
\begin{small}
\begin{tabular}{ cc}
\toprule
 {\textbf{Parameter}}& {\textbf{Value}}\\
\midrule
 $d$  & 10 \\
 $k$  & 10 \\
 $\beta_1$   & 0.1 \\
 $\beta_2$  & 100 \\
 $\beta_3$   & 0.0001 \\
 Batch Size  & 32 \\
 Optimizer & Adam \\
 Learning rate   & 0.0005 \\
 Adam: (beta1, beta2, epsilon)  & (0.9,0.99,1e-8) \\
%  Adam: beta2 &  \\
%  Adam: epsilon & 1e-8 \\
\bottomrule
\end{tabular}
\end{small}
\end{center}
\captionof{table}{Hyper-parameter settings for the experiments in Table 1-4 of the main paper\label{tab:hyperparams}.}
\end{minipage}%
\hfill
\begin{minipage}{.48\linewidth}
% \begin{figure}
    \centering
    \begin{overpic}[width=0.7\linewidth,trim=-0.8cm -0.5cm 0 -1.8cm,clip]{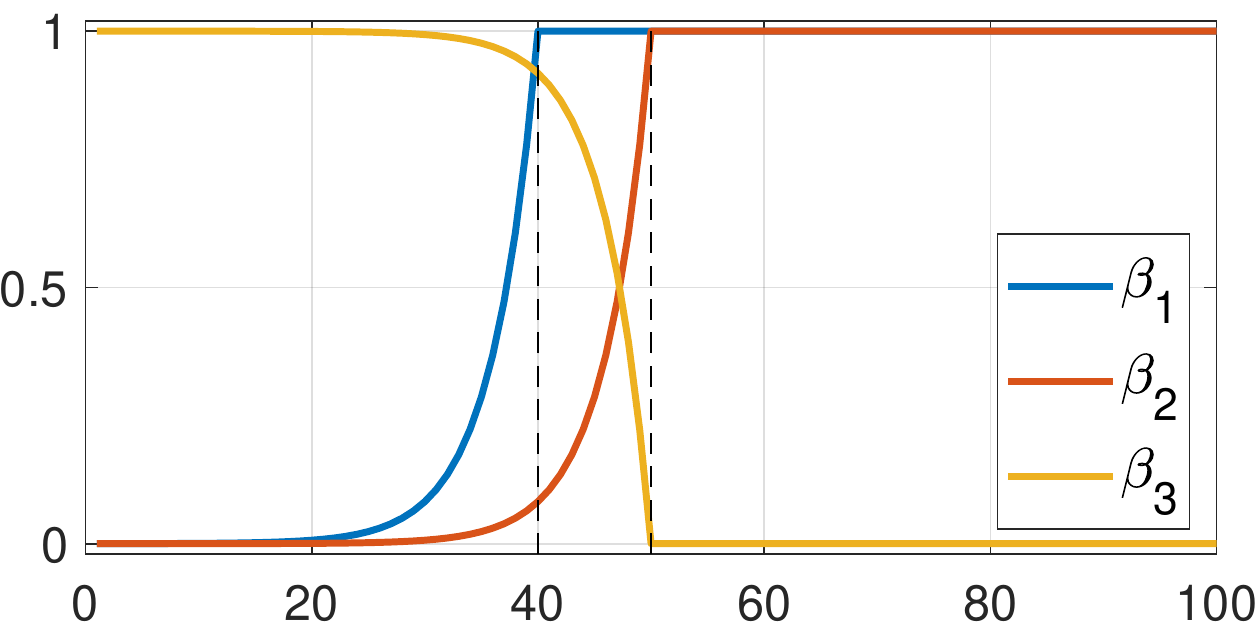}
    \put(47,0){\small epochs}
    \put(0,28){\small \rotatebox[]{90}{$\beta$ relative value}}
    \end{overpic}
    \captionof{figure}{Evolution of the regularization parameters $\beta_i, i=1,2,3$ as a function of the epoch number. Here, the parameters are all scaled to have a maximal value of one.}
    \label{fig:betas}
% \end{figure}

\end{minipage}
\end{minipage}
\begin{figure*}[t!]
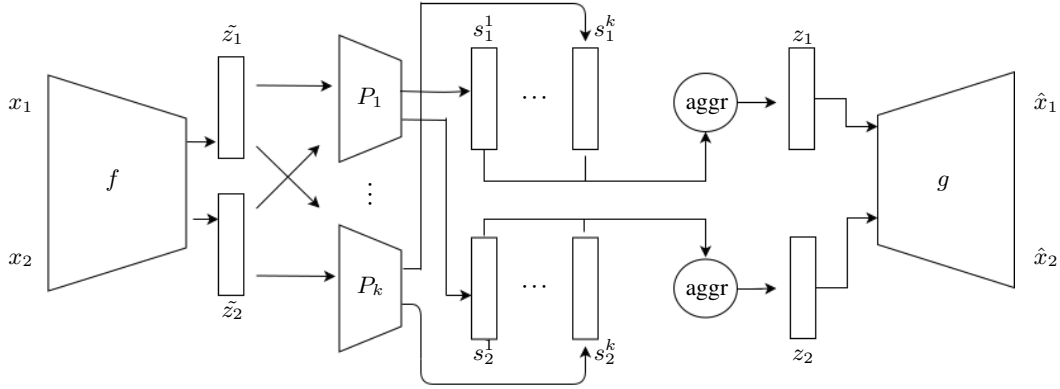

\centering
    \begin{overpic}[trim=0cm 0cm 0cm 0cm,clip,width=0.75\linewidth]{./figures/model_diag2.pdf}
     
    \put(-4, 29.5){\footnotesize $x_{1}$ }
    \put(-4, 13.5){\footnotesize $x_{2}$ }
    \put(32, 30){\footnotesize $P_{1}$ }
    \put(33, 21){$\cdot$ }
    \put(33, 20){$\cdot$ }
    \put(33, 19){$\cdot$ }

    \put(32, 10.5){\footnotesize $P_{k}$ }
    \put(6, 21.5){\footnotesize $f$ }
    \put(92, 21.5){\footnotesize $g$ }
    \put(18, 36.5){\footnotesize $\tilde{z_1}$ }
    \put(18, 8){\footnotesize $\tilde{z_2}$ }
    \put(44, 37){\footnotesize $s_1^1$ }
    \put(49, 30){$\cdot$ }
    \put(50, 30){$\cdot$ }
    \put(51, 30){$\cdot$ }
    \put(56.5, 37){\footnotesize $s_1^k$ }
    \put(44, 3.5){\footnotesize $s_2^1$ }
    \put(49, 10.5){$\cdot$ }
    \put(50, 10.5){$\cdot$ }
    \put(51, 10.5){$\cdot$ }
    \put(56.5, 3.5){\footnotesize $s_2^k$ }
    \put(77, 36.5){\footnotesize $z_1$ }
    \put(77, 3.5
    ){\footnotesize $z_2$ }
     
    \put(102, 29.5){\footnotesize $\hat{x}_{1}$ }
    \put(102, 13.5){\footnotesize $\hat{x}_{2}$ }
    \put(66.0, 29.5){\footnotesize aggr}
    \put(66.0, 10.3){\footnotesize aggr}
     
   \end{overpic}
\caption{
The architecture of our model. We process data in pairs $(x_1,x_2)$, which are embedded into an \textcolor{red}{intermediate} lower dimensional space \textcolor{red}{$\tilde{\mathcal{Z}}$}  via a siamese network $f$. The image $(z_1,z_2)$ is then mapped into $k$ smaller spaces  \textcolor{red}{$\mathcal{S}_1, \dots ,\mathcal{S}_k \subset\mathcal{Z}$} via the nonlinear operators $P_i$. The resulting vectors are aggregated in \textcolor{red}{$\mathcal{Z}$}, \textcolor{red}{with $aggr= +$}, and mapped back to the input data space by the decoder $g$. As we do not impose any constraint on $f$ and $g$, the intermediate module of the proposed architecture can be in principle attached to any autoencoder model.
\label{fig:architecture}}
\end{figure*}
\section{Training process}
We split the training process in two stages: (i) a \emph{reconstruction phase}, and (ii) a \emph{disentanglement phase}. 
This strategy helps in obtaining better results; this is  due to the fact that our distance loss $\mathcal{L}_{dis}$ needs to operate in a latent space $\mathcal{Z}$ already structured, where the distances are meaningful.
%to the fact that the reconstruction loss is in contrast with the disentanglement losses (as can be seen in Figure \ref{fig:ablation}), and having an already well-structured latent space $\mathcal{Z}$ helps the subsequent disentanglement. 
Moreover, our consistency loss makes use of the reconstructed observations, that have to be well formed to make it relevant.
We stress that the two phases are not completely separated, since the space $\mathcal{Z}$ continues to be optimized during the disentanglement phase. 

In practice, we implement this by back-propagating only through the reconstruction loss for the first $20\%$ of the training iterations. Then, the losses enter one after the other in the following order: $\mathcal{L}_{reg},\mathcal{L}_{dis},\mathcal{L}_{cons}$ in a slow-start mode. This is obtained by exponentially increasing the regularization parameters $\beta_i$ for $i=1,2$ during the training, until they reach their maximal value (as reported in Table \ref{tab:hyperparams}), with $\beta_2$ being shifted in time (number of iterations/epochs) with respect to $\beta_1$ . Conversely, we set $\beta_3=(1-\beta_2)$, so it exponentially decays until it reaches zero; indeed, the regularization loss prevents the subspace from collapsing until the other losses are active at full capacity. We show an example of the behavior of the $\beta$'s in Figure \ref{fig:betas}.

\section{Subspace structure}
\subsection{The latent subspace structure}
The model architecture, shown in Figure 3 of the main paper and reported also here in Figure \ref{fig:architecture} for convenience, imposes a factorized structure on the latent space $\tilde{\mathcal{Z}}$ into subspaces $\mathcal{S}_i ,i=1..k$. In principle, the aggregator function depicted could be any linear or nonlinear aggregation operation. In our experiments we simply choose to \textit{sum} all the subspaces, for the following reason: due to the sparsity induced on the subspaces by the loss $\mathcal{L}_{spar}$, the sum operation provides us with an approximation of the cartesian product, leading to $\tilde{\mathcal{Z}} \approx \mathcal{S}_1 \times .. \times \mathcal{S}_k $. More precisely, if the sparsity contraint holds (\textit{i.e.} $\mathcal{L}_{spar}=0$), the sum operation will be equivalent to taking the cartesian product on the latent subspace vectors, since on each dimension $r \in 1 \ldots d$ such that $\mathbf{s}_i[r] \neq 0$, for an $i \ \in 1, \ldots ,k$ the loss  $\mathcal{L}_{spar}$ enforces the latent vectors to have $\mathbf{s}_q[r] = 0 \ \  \forall q \neq i \ \in 1, \ldots ,k $. We prove this in the following:

\textbf{Sketch of proof.\hspace{0.5em}}
We prove that the sparsity imposes the structure of a product space on the latent subspace vectors. We do this by studying the first order optimality conditions for $\mathcal{L}_{spar}$, $    \frac{\partial{\mathcal{L}_{spar}(\mathbf{s}_i)}}{\partial \mathbf{s}_i}= 0   $, where  with $\mathbf{s}_i= P_i(f(x))$ we denote a latent vector in the subspace $\mathcal{S}_i$. Indicating with $\odot$ the element-wise product, we can write:

\begin{align}
    \mathcal{L}_{spar} = \sum_{i=1}^{k} \mathcal{L}_{spar}^{\mathbf{s}_i} =\sum_{i=1}^{k}\| \mathbf{s}_i \odot \sum_{j\neq i}^{k} \mathbf{s}_j\|_1 \,, \quad \mathrm{where~} \mathcal{L}_{spar}^{\mathbf{s}_i} = \| \underbrace{ \mathbf{s}_i \odot \sum_{j\neq i}^{k} \mathbf{s}_j}_Q\|_1 \,.
\end{align}

We aim to study $\frac{\partial{\mathcal{L}^{\mathbf{s}_i}_{spar}}}{\partial \mathbf{s}_i}= 0$, $\forall i \in 1, \ldots , k$ that is equivalent to: 
\begin{align}
    \frac{\partial{\mathcal{L}^{\mathbf{s}_i}_{spar}}}{\partial \mathbf{s}_i}= \frac{\partial \|Q\|_1}{\partial Q}\frac{\partial Q}{\partial \mathbf{s}_i} =  \Big(sign(\mathbf{s}_i \odot \sum_{j\neq i}^{k} \mathbf{s}_j)\Big) 
    (\sum_{j\neq i}^{k} \mathbf{s}_j) = 0, \ \forall i \in 1, \ldots , k\,.
    \label{eq:firstorder}
\end{align}
% \begin{align}
%     &\frac{\partial{\mathcal{L}^{\mathbf{s}_i}_{spar}}}{\partial \mathbf{s}_i}= 0   \ \ \forall i \in 1 \ldots k\label{eq:firstorder} \\ 
%     &\frac{\partial{\mathcal{L}^{\mathbf{s}_i}_{spar}}}{\partial \mathbf{s}_i}= \frac{\partial \|Q\|_1}{\partial Q}\frac{\partial Q}{\partial \mathbf{s}_i}\\
%     &\frac{\partial{\mathcal{L}_{spar}^{\mathbf{s}_i}}}{\partial \mathbf{s}_i}= \Big(sign(\mathbf{s}_i \odot \sum_{j\neq i}^{k} \mathbf{s}_j)\Big) 
%     (\sum_{j\neq i}^{k} \mathbf{s}_j)
% \end{align}
 W.l.o.g. we fix a dimension  $r \in 1, \ldots , d$ in the latent space. By indicating with $\mathbf{s}_i[r]$ the $r$-th entry of $\mathbf{s}_i$ we can write:
\begin{align}
    &\frac{\partial{\mathcal{L}_{spar}^{\mathbf{s}_i}}}{\partial \mathbf{s}_i}[r]= \Big(sign(\mathbf{s}_i[r]  \sum_{j\neq i}^{k} \mathbf{s}_j[r])\Big) 
    (\sum_{j\neq i}^{k} \mathbf{s}_j[r]) = 0 \,. \label{eq:fixedr}
\end{align}
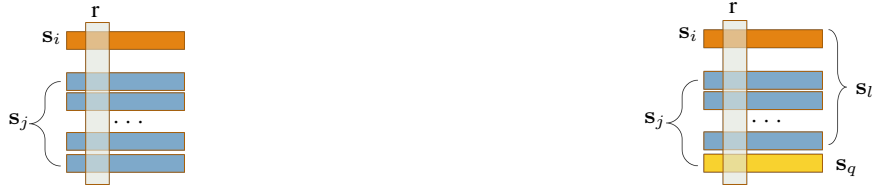
\begin{figure}
\begin{minipage}{.5\linewidth}
    \centering
    \begin{overpic}[trim=0 0cm 0 -20,clip,width=0.24\linewidth]{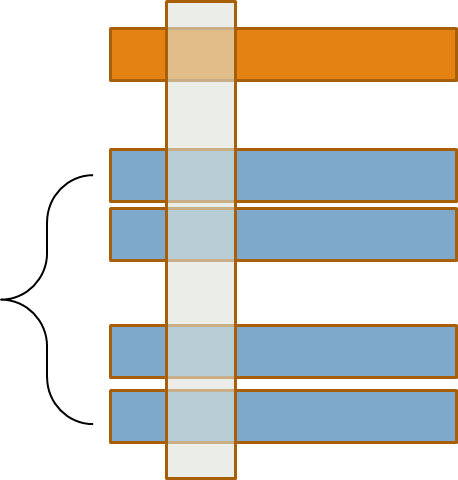}
    \put(35,95){\small r}
    \put(7,81){\small $\mathbf{s}_i$}
    \put(47,32){\footnotesize $\cdot \cdot \cdot$}
    \put(-12,34){\small $\mathbf{s}_j$}
    \end{overpic}
    %\caption{Caption}
    %\label{fig:my_label}
\end{minipage}
\begin{minipage}{.5\linewidth}
    \centering
    \begin{overpic}[trim=0 0cm 0 -20,clip,width=0.28\linewidth]{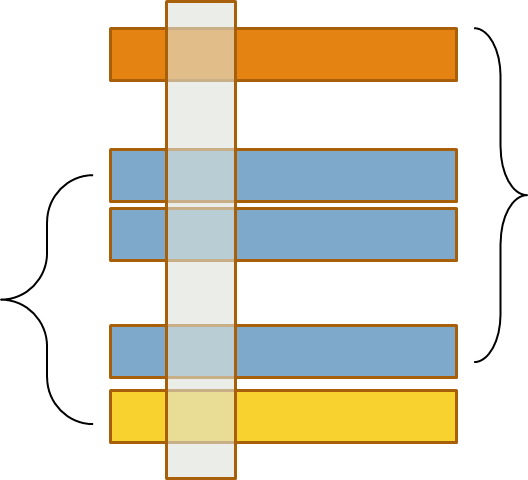}
    \put(35,95){\small r}
    \put(7,81){\small $\mathbf{s}_i$}
    \put(-12,34){\small $\mathbf{s}_j$}
    \put(47,32){\footnotesize $\cdot \cdot \cdot$}
    \put(94,8){\small $\mathbf{s}_q$}
    \put(105,50){\small $\mathbf{s}_{l}$}
    \end{overpic}
    %\caption{Caption}
\end{minipage}
\caption{\label{fig:proof} A visualization of the main variables involved in our proof.
}
\end{figure}

To satisfy Eq.~\eqref{eq:fixedr}, we have three possible cases: 

\begin{itemize}
\item \emph{Case 1:} $\mathbf{s}_i[r]=0$ and
$\sum_{j\neq i}^{k} \mathbf{s}_j[r] \neq 0$ 
\end{itemize}

\begin{itemize}
\item \emph{Case 2:} $\mathbf{s}_i[r] \neq 0$ and $  \sum_{j\neq i}^{k} \mathbf{s}_j[r]=0$  \label{eq:case2}\\ 
Since we are optimizing $\forall i $ we can consider $\frac{\partial{\mathcal{L}_{spar}^{\mathbf{s}_q}}}{\partial \mathbf{s}_q}[r] = 0$ for every other $q \in 1 \ldots k, q \neq i$, %$\sum_{j'\neq q}^{k} \mathbf{s}_{j'}[r]=0$.
Therefore we have:
\begin{align}
    %&\frac{\partial{\mathcal{L}_{spar}^{\mathbf{s}_q}}}{\partial \mathbf{s}_q}= \Big(sign(\mathbf{s}_q \odot \sum_{l\neq q}^{k} \mathbf{s}_l)\Big) (\sum_{l\neq q}^{k} \mathbf{s}_l)\\   
    &\frac{\partial{\mathcal{L}_{spar}^{\mathbf{s}_q}}}{\partial \mathbf{s}_q}[r]= \Big(sign(\mathbf{s}_q[r]  \sum_{l\neq q}^{k} \mathbf{s}_l[r])\Big) 
    (\sum_{l\neq q}^{k} \mathbf{s}_l[r]) = 0
    \label{eq:Lsq}
\end{align}
%if:
%\begin{align}
%    &\sum_{j\neq i}^{k} \mathbf{s}_j[r]=0 \wedge \mathbf{s}_i[r]\neq 0
%\end{align}
We can split this latter case in two subcases:
\begin{itemize}
\item \emph{Case 2.1:} $\mathbf{s}_q[r]=0$, $\forall q \neq i$ \\
Therefore, satisfying our thesis.

\item \emph{Case 2.2:} $\mathbf{s}_q[r]\neq0$ for at least one $q\neq i$.

In this situation, we can write:
\begin{align}
\sum_{j'\neq q,i}^{k} \mathbf{s}_{j'}[r] + \mathbf{s}_{q}[r] = 0 \mbox{ and thus }  \sum_{j'\neq q,i}^{k} \mathbf{s}_{j'}[r]=-\mathbf{s}_{q}[r] .
\end{align}
From which we have:
\begin{align}
    \sum_{j'\neq q}^{k} \mathbf{s}_{j'}[r]= \mathbf{s}_{i}[r] + \sum_{j'\neq q,i}^{k} \mathbf{s}_{j'}[r] = \mathbf{s}_{i}[r]-\mathbf{s}_{q}[r].
\end{align} 

Substituting in Eq.\ref{eq:Lsq} (by replacing $j'$ with $l$) we get:
\begin{align}
    &\frac{\partial{\mathcal{L}_{spar}^{\mathbf{s}_q}}}{\partial \mathbf{s}_q}[r]= \Big(sign(\mathbf{s}_q[r]  (\mathbf{s}_i[r]-\mathbf{s}_q[r]))\Big) 
    (\mathbf{s}_i[r]-\mathbf{s}_q[r])
\end{align}
Now because we have that $\mathbf{s}_q [r] \neq 0$, this implies:
\begin{align}
    &\frac{\partial{\mathcal{L}_{spar}^{\mathbf{s}_q}}}{\partial \mathbf{s}_q}[r] =0 \iff \mathbf{s}_i[r] -\mathbf{s}_q[r]=0 \implies \mathbf{s}_i[r] =\mathbf{s}_q[r] %\vee \mathbf{s}_q[r]=0
\end{align}
and this holds $\forall q \in 1, \ldots , k $ and $q\neq i $ such that $\mathbf{s}_q [r] \neq 0$.
(referring to Figure \ref{fig:proof} may help the reader). 

This allows us to conclude that: $\sum_{j\neq i}^{k} \mathbf{s}_j[r]= \alpha \mathbf{s}_i[r] \neq 0$, with $\alpha$ being an integer between $1$ and $k-1$.
Therefore we get a contradiction with our hypothesis of \emph{Case 2} $\mathbf{s}_i[r] \neq 0$ and $  \sum_{j\neq i}^{k} \mathbf{s}_j[r]=0$, and thus the unique possible subcase is the former \emph{Case 2.1}.
%Therefore we get a contradiction with our hypothesis $\mathbf{s}_q[r]\neq0$ and $\mathbf{s}_i[r]\neq 0$, and the unique possible case is the former.
%We can repeat the same reasoning for all other $q'\neq i,q \in 1 \ldots k$
\end{itemize}
\end{itemize}

\begin{itemize}
\item \emph{Case 3:} $\mathbf{s}_i[r]=0$ and $  \sum_{j\neq i}^{k}
\mathbf{s}_j[r]=0$ \\
Performing the same analysis done in \emph{Case 2}, in this case we get that $\forall i \in 1 \ldots k  \  \mathbf{s}_i[r]=0$. Therefore, all the latent subspace vectors will have the same dimension $r$ set to zero.  In this case, we can consider recursively the other $k-1$ dimensions. The case where all dimensions $r \in 1 \ldots k$ are zero, for all $\mathbf{s}_i, i=1 \ldots k $ is theoretically possible, but we stress this is rather an exotic case that cannot happen in practice, as we comment in the last paragraph below.
Since we have chosen $r$ w.l.o.g., the same s true for all dimensions in $1 \ldots d$.
%Then in the former case we would have that on dimension $r$ the vectors $\mathbf{s}_j$ in the subspaces $\mathcal{S}_j, \ j\neq i$ will sum to zero and in the latter we would have that $r$-th entry on the vector belonging to the subspace $\mathcal{S}_i$ is zero.
%Since we optimize $\mathcal{L}_{spar}(\mathbf{s}_i) \ \forall i \in 1...k$, optimizing for $\| \mathbf{s}_i \otimes \sum_{j\neq i}^{k} \mathbf{s}_j\|_1 $ is equivalent constraint all the possible $\binom{k}{2}$ pairs of vector to be orthogonal and sparse, i.e. $\| \mathbf{s}_i \otimes  \mathbf{s}_j\|_1  \forall i,j  \in 1 \ldots k, \ s.t. i \neq j$. \\
Therefore, we have that each vector $\mathbf{s}_i$ will be nonzero in the $l>0$ dimensions where the other $\mathbf{s}_j$ are zero.
Now setting $aggr=+$, we have that the sum corresponds to concatenating the latent subspace vectors along the nonzero dimensions, i.e. taking the cartesian product of the subspace to get an element of $\tilde{\mathcal{Z}}$.
\end{itemize}

\textbf{Degenerate case} In the proof we mentioned the degenerate case in which $\mathbf{s}_i[r]=0\; \forall i \in 1 \ldots k, \ \forall r \in 1 \ldots d$. This would mean that the latent subspaces have collapsed to the same point (a vector made of zeros). This exotic case is never reached in practice, due to the other losses such as the reconstruction loss, the consistency losses, and the contrastive term of the distance loss.
% \begin{table}[h!]
% \caption{Average standard deviation on the latent subspaces in Figure \ref{fig:latent_space}}
% \label{tab:conv}

% \begin{center}
% \begin{small}
% \begin{tabular}{ cc}
% \toprule
%  {\textbf{Subspace}}& {\textbf{Average std}}\\
% \midrule
%  $\mathcal{S}_1$ & 0.061 \\
%  $\mathcal{S}_2$ & 0.043 \\
%  $\mathcal{S}_3$ & 0.063 \\
%  $\mathcal{S}_4$ & 0.026 \\
%  $\mathcal{S}_5$ & 2.6e-5 \\
%  $\mathcal{S}_6$ & 2.1e-5 \\
%  $\mathcal{S}_7$ & 1.5e-5 \\
%  $\mathcal{S}_8$ & 0.058 \\
%  $\mathcal{S}_9$ & 0.063 \\
%  $\mathcal{S}_{10}$ & 2.1e-5 \\
% \bottomrule
% \end{tabular}

% \end{small}
% \end{center}
% \end{table}

\end{fleqn}

\end{document}